\documentclass{article}

\usepackage{arxiv}

\usepackage[utf8]{inputenc} 
\usepackage{hyperref}       
\usepackage{url}            
\usepackage{booktabs}       
\usepackage{amsfonts}       
\usepackage{nicefrac}       
\usepackage{microtype}      

\usepackage{graphicx}
\usepackage{multirow}
\usepackage{amsmath}

\usepackage{subfig}

\usepackage{import}

\renewcommand{\vec}[1]{\boldsymbol{#1}}     
\newcommand{\vX}{{\vec{X}}}

\DeclareMathOperator{\E}{E}

\newcommand{\euler}{\mathrm{e}}             

\DeclareMathOperator{\cov}{cov}
\DeclareMathOperator{\var}{var}

\title{Missing Features Reconstruction and Its Impact on Classification Accuracy}

\author{
  Magda Friedjungová \\
  Faculty of Information Technology\\
  Czech Technical University in Prague\\
  Prague, Czech Republic\\
  \texttt{magda.friedjungova@fit.cvut.cz} \\
   \And
  Daniel Vašata \\
  Faculty of Information Technology\\
  Czech Technical University in Prague\\
  Prague, Czech Republic\\
  \texttt{daniel.vasata@fit.cvut.cz} \\
  \And
  Marcel Jiřina \\
  Faculty of Information Technology\\
  Czech Technical University in Prague\\
  Prague, Czech Republic\\
  \texttt{marcel.jirina@fit.cvut.cz} \\
}

\begin{document}
\maketitle

\begin{abstract}
  In real-world applications, we can encounter situations when a well-trained model has to be used to predict from a damaged dataset. The damage caused by missing or corrupted values
  can be either on the level of individual instances or on the level of entire features.
  Both situations have a negative impact on the usability of the model on such a dataset.
  This paper focuses on the scenario where entire features are missing which can be understood as a specific case of transfer learning. Our aim is to experimentally research
  the influence of various imputation methods on the performance of several
  classification models.
  The imputation impact is researched on a combination of traditional methods such as $k$-NN,
  linear regression, and MICE compared to modern imputation methods such as
  multi-layer perceptron (MLP) and gradient boosted trees (XGBT).
  For linear regression, MLP, and XGBT we also propose two approaches to using them for multiple features imputation.
  The experiments were performed on both real world and
  artificial datasets with continuous features where different numbers of features,
  varying from one feature to $50\%$, were missing.
  The results show that MICE and linear regression
  are generally good imputers regardless of the conditions. On the other hand,
  the performance of MLP and XGBT is strongly dataset dependent.
  Their performance is the best in some cases, but more often they perform worse
  than MICE or linear regression.
\end{abstract}


\label{sec:intro}
While solving a classification task one often faces demanding preprocessing of data.
One of the preprocessing steps is the treatment of missing values. In practice, we struggle with randomly
located single missing data in instances or with missing entire features.
In real-world scenarios, e.g. \cite{zhang2013,zhu2015,jordanov2018}, we have to deal with missing data. Missing values can also be part of a cold-start problem.
Imputation treatments for missing values have been widely investigated
\cite{little2014,buuren2018,alireza08}
and plenty of methods how to reconstruct missing data were designed, but these methods are not directly
designated for entire missing features reconstruction.

This work focuses on the influence of missing entire features and possibilities of their
reconstruction for usage in predictive modeling.
We consider the following scenario: a classification model is trained on a dataset containing
a complete set of continuous features but has to be used for prediction of classes of a dataset with some
entire features missing. Entire feature reconstruction and its usage in an already learned model
in order to perform with a reconstructed dataset distinguishes our work from others.
Our point of interest is to find out how missing features impact the accuracy
of the classification model, what possibilities of missing entire
features reconstruction exist, and how the model performs with imputed data.
In our work, the reconstruction of missing features, i.e. data imputation,
is the very first task of transfer learning methods~\cite{Pan10}, where the identification
of identical, missing, and new features is crucial.

Experimental results of this work should shed more light onto the applicability
of state-of-the-art imputation methods on data
and their ability to reconstruct entire missing features.
We deal with traditional imputation methods: linear regression, $k$-nearest neighbors ($k$-NN), and multiple imputation by chained equations (MICE) \cite{buuren2018},
as well as with modern methods: multi-layer perceptron (MLP),
and gradient boosted trees (XGBT) \cite{chen16}.
Experiments are performed on four real and six artificial datasets.
The imputation influence is studied on six commonly used binary classification models: random forest,
logistic regression, $k$-NN, naive Bayes, MLP, and XGBT.
The amount of missing data varies between one feature and $50\%$.

This paper is structured as follows.
In the next section we briefly review related work.
Section \ref{sec:imp_methods}
introduces imputation methods that are being analyzed in this work.
Multiple features imputation is also discussed here.
In Section \ref{sec:experiments} we describe the experiments that were carried out and present their results in
Section \ref{sec:experimental_results}. Finally, we conclude the paper in Section \ref{sec:conclusion}.

\section{Related Work}
\label{sec:relatedwork}
There exist many surveys which summarize missing value imputation methods such as \cite{alireza08,zhang2013,baitharu2013,silva2015,murray2018,jordanov2018,junninen04}.
A lot of them are more than five years old and focus on traditional imputation methods.

A very good review of methods for imputation of missing values was provided by \cite{alireza08}. This study is focused on discrete values only with up to $50\%$ missingness. They experimentally evaluated six imputation methods (hot-deck, imputation framework with hot-deck, naive Bayes, imputation framework with naive Bayes, polynomial multiple regression, and mean imputation) on 15 datasets used in 6 classifiers. Their results show that all imputation methods except for mean imputation improve classification error when missingness is more than $10\%$. The decision tree and naive Bayes classifiers were found to be missing data resistant, however other classifiers benefit from the imputed data.

In \cite{zhang2013}, performance of imputation methods was evaluated on datasets with varying amounts of missingness (up to $50\%$). Two scenarios were tested: values are missing only during the prediction phase, and values are missing during both induction and prediction phases. Three classifiers were used in this study: a decision tree, $k$-NN, and a Bayesian network. Imputation by mean, $k$-NN, regression and ensemble were used as imputation methods. The experimental results show that the presence of missing values always leads to performance reduction of the classifier, no matter which imputation method is used to deal with the missing values. However, if there are no missing data in the training phase, imputation methods are highly recommended at the time of prediction.

Finally, in \cite{Arroyo2018},
W. Arroyo et al. present imputation of missing values of Ozone in real-life datasets using various imputation methods - multiple linear and nonlinear regression, MLP and radial basis function networks, where the usefulness of artificial neural networks is presented.

\section{Imputation Methods}
\label{sec:imp_methods}
Plenty of methods of missing data reconstruction have been designed.
They perform differently on various datasets and in practice the most suitable
imputation method for a given dataset is usually chosen according to the evaluation
of the average performance (e.g. RMSE) of each method in the phase of training \cite{Salgado2016}.

First let us briefly introduce imputation methods which we focus on within this study.
The most basic methods are linear regression and the $k$-NN (see e.g. \cite{jonsson04}).

The MICE \cite{Schafer1997,buuren2018} does not simply
impute missing values using the most fitting single value, but
it also tries to preserve some of the randomness of the original data distribution.
This is being accomplished by performing multiple imputations, see \cite{rubin87}.
The MICE comes up with very good results and is currently one of the best-performing methods \cite{buuren2018}.
In our research we use MICE in a simplified way. This means that multiple imputations are pooled using
the mean before the classification model is applied. The reason is that we want to simulate the situation when
the use of a classification model is restricted.

The MLP \cite{silva2015} with at least one hidden layer and no activation function
in the output layer and the XGBT, see~\cite{chen16} for more details,
are considered to be modern imputation methods.

\subsection{Multiple Features Imputation}
\label{subsec:multiple_imp_methods}
To impute several missing features, there are two ways of accomplishing this task using the previously
mentioned methods. The first is to impute all features simultaneously which can be done using $k$-NN and MLP models.
The second, which is usable for all other methods, is to apply the model sequentially
one missing feature after another. However, to do this, it is important to choose some
order in which the features will be imputed. We focus on an ordering where
the most easy to impute features are treated first.

In the case of $k$-NN and MICE such a sequential imputation is not needed. The reason is that, in the case of $k$-NN, the neighbors
typically do not change in subsequent steps and MICE is already prepared for multiple features imputation
using an internal chained equation approach \cite{Schafer1997,buuren2018}.

\subsection*{Linear Imputability}
A simple way of measuring imputation easiness of features is to use the multiple correlation
coefficient~\cite{anderson}. Multiple correlation coefficient $\rho_ {X, \vX'}$
between a random variable $X$ and a random vector
$\vX' = (X_1', \dotsc, X_n')^T$
is the highest correlation coefficient between $X$ and a linear combination
$\alpha_1 X_1' + \dotsc + \alpha_n X_n' = \vec \alpha^T \vX'$
of random variables $X_1', \dotsc, X_n'$,
\[
  \rho_ {X, \vX'} = \max_{\vec \alpha \in \mathbb{R}^n}
  \rho_{X, \vec \alpha^T\vX'}.
\]

It takes values between $0$ and $1$, where $\rho_ {X, \vX'} = 1$
means that the prediction by linear regression of $X$ based on $\vX'$
can be done perfectly and $\rho_ {X, \vX'} = 0$ means that the linear regression
will not be successful at all.

When $X_1,\dotsc, X_p$ are the $p$ features, we call the multiple correlation coefficient
$\rho_{X_i, \vX_{-(i)}}$
between $X_i$ and a random vector of other features
$\vX_{-(i)} = (X_1, \dotsc, X_{i-1}, X_{i+1}, \dotsc, X_p)^T$
the \emph{linear imputability} of feature $X_i$.

The estimation of the linear imputability is based on the following expression
\[
  \rho_{X_i, \vX_{-(i)}}^2 = \frac{\cov(X_i, \vX_{-(i)})^T \big(\cov(\vX_{-(i)})\big)^{-1}
  \cov(X_i, \vX_{-(i)})}{\var(X_i)},
\]
where $\cov(X_i, \vX_{-(i)})$
is a vector of covariances between $X_i$ and remaining features $X_1, \dotsc, X_{i-1}, X_{i+1}, \dotsc, X_p$, and
$\cov(\vX_{-(i)})$ is a $p-1 \times p-1$ variance-covariance matrix of covariances between
remaining features.

If we want to impute multiple features, say $X_i, X_{i+1}, \dotsc, X_{i+k}$,
in the first step we choose $X_j, i \leq j \leq i+k$ such that
$\rho_{X_j, \vX_{-(i, \dotsc, i+k)}}$ is the largest,
where $\vX_{-(i, \dotsc, i+k)} = (X_1, \dotsc, X_{i-1}, X_{i+k+1}, \dotsc, X_p)^T$
is a vector of the remaining features.
Then, in the next step, we repeat the process where $X_j$ is taken as a known feature.
Thus we choose $X_l, i \leq l \leq i+k, l \neq j$ such that its linear imputability with respect
to random vector $\vX_{-(i, \dotsc, j-1, j+1, \dotsc, i+k)}$ is the largest.
We continue this way until all missing features are imputed.

Note that we are recalculating linear imputability in every step.
This should not be done if the imputation is performed with linear regression since after
the re-estimation (on the full training set) one obtains unachievable values.

\subsection*{Information Imputability}
Linear imputability is a simple measure of how the linear regression imputation will perform.
However, when one uses more sophisticated imputation models like MLP or XGBT that can handle
non-linear dependencies, the linear imputability may not be suitable.

Hence we propose another way how to measure the imputability which is based on a particular result
from Information theory.
If a feature $X_j$ is predicted by an estimator $\hat X_j$ based on other features represented by a vector
$\vX_{-(j)}$,
i.e. $\hat X_j \equiv \hat X_j\big(\vX_{-(j)}\big)$, then it can be shown (see \cite{CoverThomasInformation}) that
\[
  \E\big(X_j - \hat X_j\big)^2 \geq \frac{1}{2\pi \euler} \euler^{2H(X_j | \vX_{-(j)})},
\]
where $H(X_j | \vX_{-(j)})$ is the conditional (differential) entropy
of $X_j$ given $\vX_{-(j)}$.

Hence the lower bound of the expected prediction error is determined by the conditional entropy
$H(X_j | \vX_{-(j)})$. The greater the entropy is the worse predictions one can achieve
at best when estimating $X_j$ from other features.
Thus one may measure imputability through the value of a conditional entropy multiplied by $-1$ in order
to have larger values which correspond to better imputability.
Hence we define the \emph{information imputability} as a value of
$-H(X_j | \vX_{-(j)})$.

The process of multiple feature imputation is now exactly the same as it was
using linear imputability. One first imputes the feature with the largest
information imputability. The only difference is that in the second and all subsequent steps
the recalculation does not make sense since one is not able to get any new information
no matter what model will be used for the imputation. This partially simplifies the process of imputation order selection.

On the other hand, the problem that strongly limits its practical usage is the estimation of the conditional entropy.
Even the most recently proposed estimators in \cite{Lombardi2016,Sricharan2013}
suffer from the curse of dimensionality.
This is due to the fact that all these estimators are based on the $k$-NN approach introduced
by Kozachenko and Leonenko in \cite{KozachenkoLeonenko1987}.
As our numerical experiments indicate, the method is limited to approximately five features depending
on the underlying joint distribution.

\section{Experiments}
\label{sec:experiments}
Our experiments consist of the following steps.
First the original dataset is divided into a training part ($70\%$) and a test part ($30\%$).
Several classification models as well as all imputation methods are trained on the training part.
The imputation models are trained to impute in scenarios where each individual feature is missing
and where randomly selected combinations of multiple features are missing.
The degree of missingness varies from $10\%$ to $50\%$.
Finally, an evaluation of the accuracy of all classification models combined with all
imputation methods is performed on the test dataset.

\subsection{Settings and Parameters of Imputation Methods}
Experiments were done using various settings.
In order to keep the report short we present only those with satisfying results.
All experiments were implemented in \texttt{Python 3}.

The $k$-NN imputation (\textit{knn}) was
implemented using the \texttt{fancyimpute} library\footnote{\texttt{Fancyimpute} repository: \url{https://github.com/iskandr/fancyimpute}}.
A missing value is imputed by sample mean of the values of its neighbors
weighted proportionally to their distances. In the case where multiple features are missing we impute all missing values at once (per row).
In the presented results the hyper-parameter $k$ is always taken as $k=5$. This value was chosen
based on preliminary experiments and with respect to computational time.

For the MICE method (\textit{mice}) we also used the \texttt{fancyimpute} library.
The parameter setup was inspired by \cite{azur2011} and we chose the number of imputations to be
$150$, the internal imputation model to be a Bayesian ridge regression, and the multiple imputed values to be pooled using the mean.

Linear regression imputation was implemented using the \texttt{scikit-learn} library\footnote{\texttt{Scikit-learn} repository: \url{https://github.com/scikit-learn/scikit-learn}} \cite{scikit}.
We tested two scenarios within the case when multiple features were missing. The first scenario was based on the
linear imputability (\textit{linreg-li})
and an iterative approach (\textit{linreg-iter})
which corresponds to chained equations in MICE. This approach repeats two steps.
First, every single missing value is imputed from the known features only. Second, all the imputed values are iteratively
re-imputed from other features (all features except the one being imputed).

The MLP imputation is implemented using the \texttt{scikit-learn} library in two scenarios. The first (\textit{mlp}) imputes all missing features at once and the second (\textit{mlp-li}) imputes subsequently based on linear imputability. The hyper-parameters of MLP (learning rate, numbers and sizes of hidden layers, activation function, number of training epochs) were tuned using randomized search.
The XGBT was implemented using the \texttt{xgboost} library\footnote{\texttt{XGBoost} repository: \url{https://github.com/dmlc/xgboost}} in two scenarios. The first (\textit{xgb-li}) is an analogy to \textit{mlp-li} and the second (\textit{xgb-iter}) to \textit{linreg-iter}. The hyper-parameters (learning rate, number of estimators, max depth of trees) were again tuned using randomized search.

The multiple features subsequent imputation scenario using information imputability is
not presented here since in preliminary experiments it does not bring any benefits over linear imputability.

\subsection{Evaluation}
Imputation methods were evaluated using six binary classification models: $k$-NN, MLP,
logistic regression (LR), XGBT, random forest (RF), and naive Bayes (NB), where LR, RF, and NB
were provided by the \texttt{scikit-learn} library.
We again used the randomized search algorithm to get classifier hyper-parameter configurations for each dataset.

First, we trained all classification models and measured their performance on the full test dataset (no missing features)
(see Table \ref{tab:datasets} for results). Second, we combined them with imputation methods.
We then measured the accuracies of all classification models on the imputed test dataset.
Finally, we calculated the imputation performances as changes with respect to the accuracies on the full test dataset.

\subsection{Datasets}
We use both artificial and real datasets which are presented in Table \ref{tab:datasets}.
All datasets have continuous features and binary target labels.
All datasets contain complete data without missing values. We assume all features are in a suitable form for the classification of the target label.

The real Wine Quality dataset originally contains ten target classes that were symmetrically
merged in order to have a binary classification task.
The artificial datasets were generated using the \texttt{make\_classification} method in the \texttt{scikit-learn} library.
They contain informative and redundant features.
Informative features are drawn independently from the standard normal distribution.
Redundant features are generated as random linear combinations of the informative features.
A noise drawn from a centered normal distribution with variance $0.1$ is added to each feature.

\begin{table}[t]
\caption{Details of datasets with corresponding classification model accuracies.
The number of features (\# feat.) does not include the target label. The name
ds\_a\_b\_c stands for an artificial dataset where $a$ is the number of features, $b$ is the number of informative features, and $c$ is the number of redundant features. Bold values of accuracy correspond to the two best models for a given dataset.}
\label{tab:datasets}
\scriptsize
\begin{center}
\begin{tabular}{|c|c|c|c|c|c|c|c|c|c|c|}
\hline
Name & Type & \# feat. & \# records & LR & MLP & $k$-NN & NB & XGBT & RF \\
\hline
Cancer \cite{uci-ml} &  real & 9 & 699 & \textbf{0.966} & 0.956 & 0.961 & 0.941 & 0.961 & \textbf{0.966} \\
MAGIC \cite{uci-ml} &  real & 10 & 19020 & 0.789 & 0.829 & 0.825 & 0.726 & \textbf{0.869} & \textbf{0.857} \\
Wine \cite{uci-ml} &  real & 11 & 4898 & 0.751 & 0.682 & 0.696 & 0.696 & \textbf{0.786} & \textbf{0.773} \\
Spambase \cite{alcala11} &  real & 57 & 4597 & 0.912 & \textbf{0.943} & 0.783 & 0.834 & 0.930 & \textbf{0.932} \\
\hline
Ringnorm \cite{alcala11} &  artificial & 20 & 7400 & 0.762 & 0.817 & 0.679 & \textbf{0.979} & \textbf{0.945} & 0.936 \\
Twonorm \cite{alcala11} &  artificial & 20 & 7400 & \textbf{0.980} & 0.978 & 0.968 & 0.980 & \textbf{0.980} & 0.952 \\
ds\_10\_7\_3 &  artificial & 10 & 4000 & 0.836 & \textbf{0.990} & 0.971 & 0.869 & \textbf{0.980} & 0.973 \\
ds\_20\_14\_6 &  artificial & 20 & 4000 & 0.839 & \textbf{0.992} & 0.959 & 0.818 & 0.968 & \textbf{0.977} \\
ds\_50\_35\_15 &  artificial & 50 & 4000 & 0.886 & \textbf{0.995} & 0.943 & 0.861 & \textbf{0.975} & 0.919 \\
ds\_100\_70\_30 &  artificial & 100 & 4000 & 0.819 & \textbf{0.975} & \textbf{0.878} & 0.790 & 0.819 & 0.792 \\
\hline
\end{tabular}
\end{center}
\end{table}

\section{Results of Experiments}
\label{sec:experimental_results}
\begin{figure}[ht]%
    \centering
    \subfloat[LR model on Twonorm dataset]{{\includegraphics[width=0.45\textwidth]{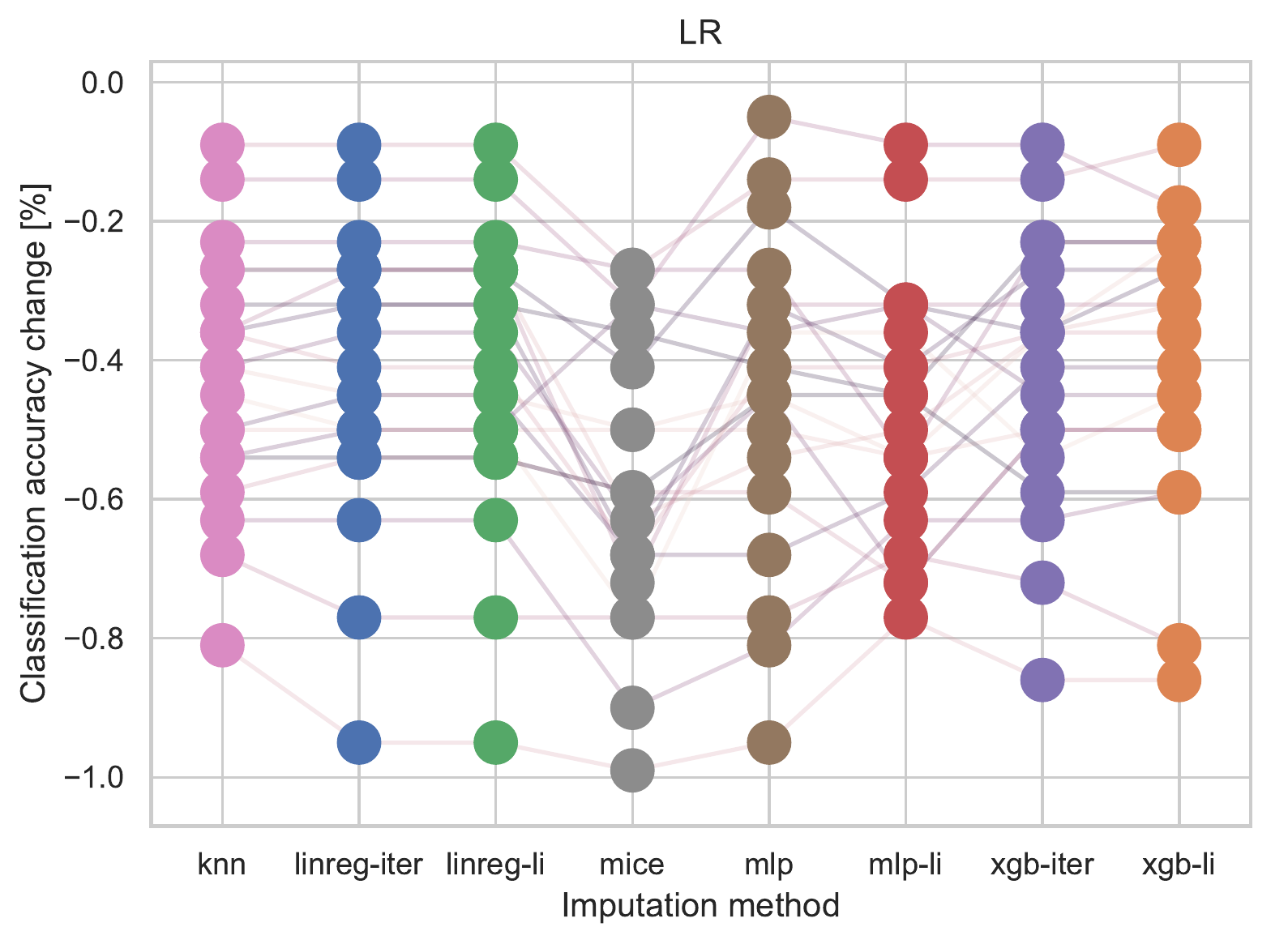} }}%
    \quad
    \subfloat[XGBT model on MAGIC dataset]{{\includegraphics[width=0.45\textwidth]{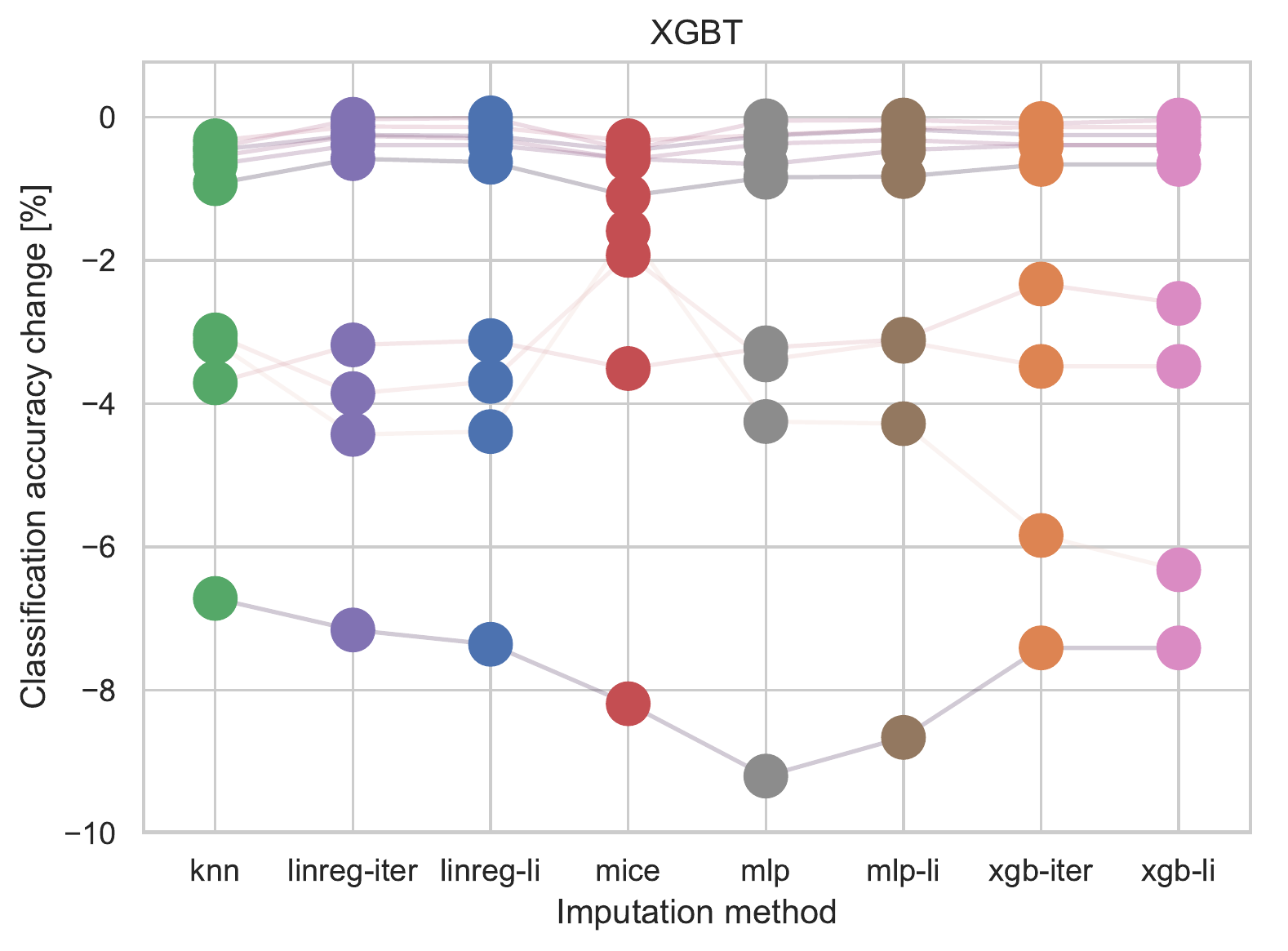} }}%
    \caption{Change in classification accuracy under all imputation methods for single missing features. Each feature is linked between different methods using a line.}%
    \label{fig:one_feature}%
\end{figure}

Results of the single feature imputation are shown in Table \ref{tab:single_feature},
where we present measured accuracy changes using the sample mean $\pm$ the sample standard deviation.
The top $10\%$ of imputation methods for each dataset and classification model are
indicated by the value printed in bold.
Two typical scenarios are shown in more details in Figure \ref{fig:one_feature}.

\begin{figure}[ht]
\centering
\includegraphics[width=0.97\textwidth]{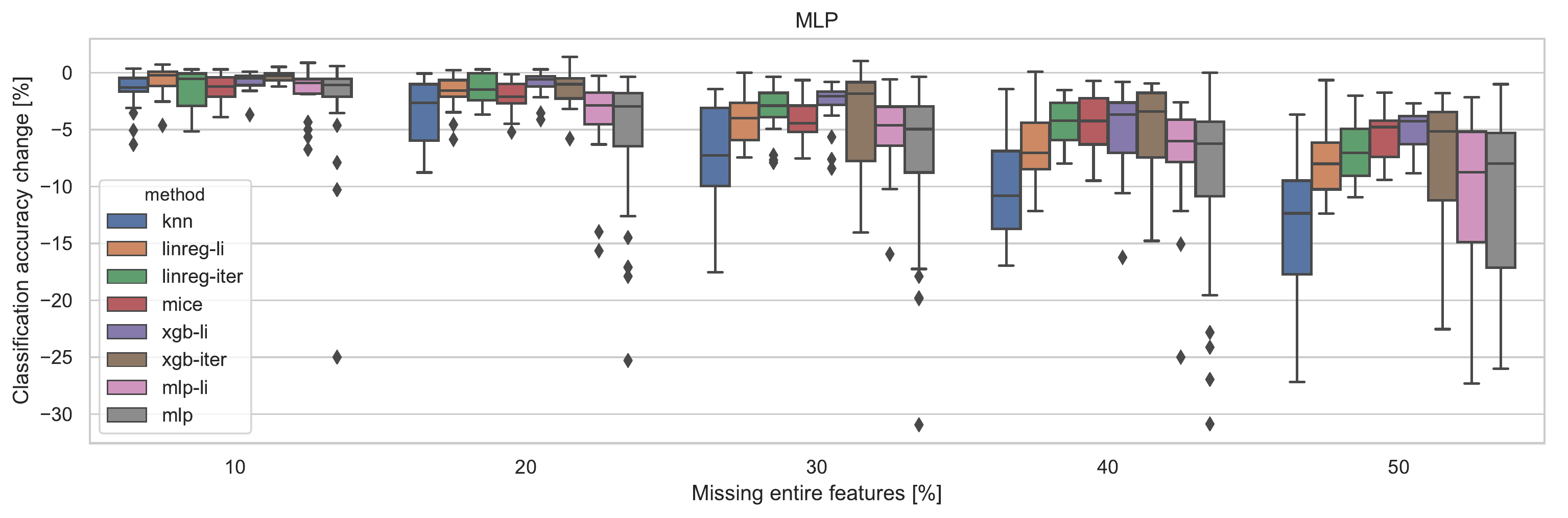}
\caption{Classification accuracy change of MLP model on real dataset Spambase.}
\label{fig:multiple_real}
\end{figure}

\begin{figure}[ht]
\centering
\includegraphics[width=0.97\textwidth]{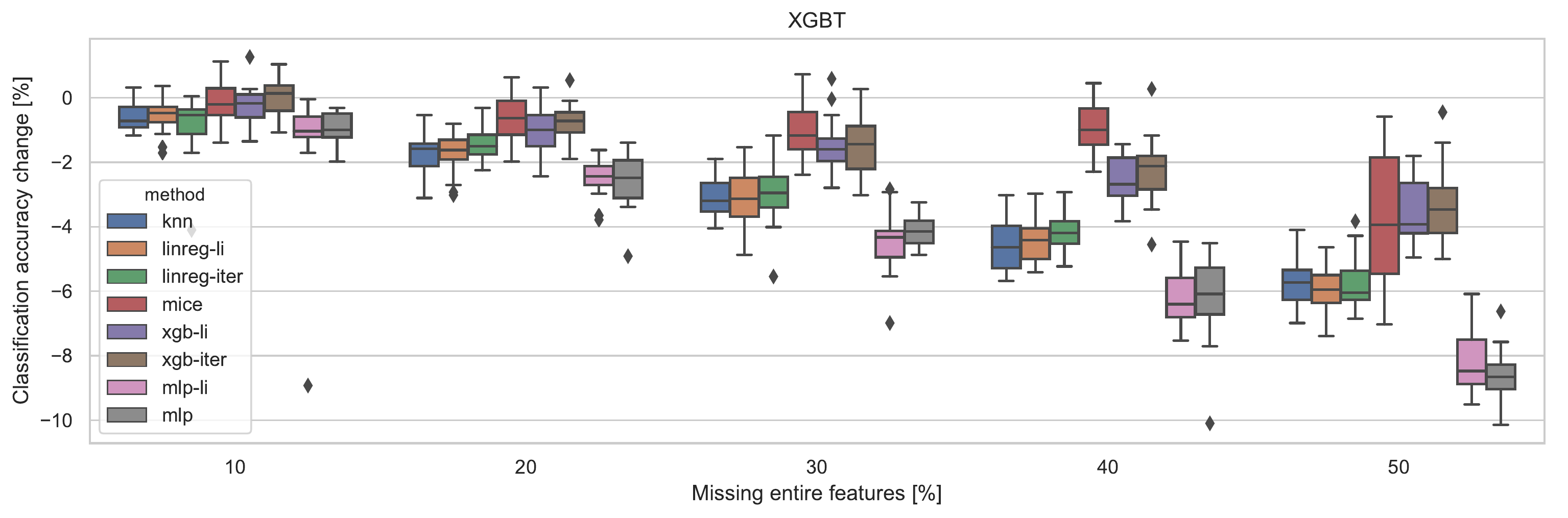}
\caption{Classification accuracy change of XGBT model on artificial dataset Ringnorm.}
\label{fig:multiple_artificial}
\end{figure}

Results of the multiple features imputation for two best models on each dataset
are presented in Tables \ref{tab:multiple_real} and \ref{tab:multiple_artificial} for real and artificial datasets, respectively.
Visualizations of typical results are given in Figure \ref{fig:multiple_real} for a selected real dataset and in Figure \ref{fig:multiple_artificial} for a selected artificial dataset. Box plots are used to show the results for different
imputation methods and portions of missing features.

\subsection{Discussion}
The results are highly dataset specific. For some datasets (Cancer, all \mbox{ds\_...} datasets)
the decreases in the classification accuracy were only minor, less than $1\%$, even for $50\%$
of missing values. On the other hand for some datasets (MAGIC, Ringnorm) the decrease is much greater, $1\% - 2\%$ for $10\%$ of missing features and $10\%$ for $50\%$ of missing features.

From the imputation methods point of view MICE usually performs the best on real datasets.
On artificial datasets it places among the best methods only for the Ringnorm and ds\_10\_7\_3
datasets. Its results often have smaller variance than results of other methods.

Results comparative to MICE were often reached using linear regression imputation
(specifically \textit{linreg-li}). Especially on artificial datasets
it usually performs the best.
In most cases either the MICE or linear regression are the best methods.

XGBT and MLP performances are much more dataset dependent. However, their performance is usually
not comparable to the best method and it also strongly depends on what classification model is used and how many features are missing. See e.g. MAGIC dataset
where MLP is performing well for $30\%$ of missing features and performing badly for $10\%$
of missing features or the Spambase where a similar discrepancy holds for XGBT.
Finally, the $k$-NN almost always performs worse than other methods.
The only exception is the ds\_20\_14\_6 dataset with random forest classification model.

Considering the amount of missing features it seems that
results depend on the portion of missing features and not on the absolute number of missing features.

When we restrict ourselves to one missing feature reconstruction,
the results are again highly dataset specific. For Cancer, Spambase, and ds\_... datasets the accuracy
after imputation actually increases. This is probably due to the fact that original classification
models were overfitted and the proper imputation enables them to generalize better.
On the other hand for the MAGIC dataset the performance decrease was around $1\% - 2\%$.

One can summarize that the best imputation methods were MICE, which performs well on real datasets
and linear regression, which performs well also on artificial datasets.
In some cases comparable results were reached by XGBT and MLP imputation.
Again, only the $k$-NN imputation is not performing well enough.

If one analyzes all classification models (not just the two best), then
classification models with higher accuracy perform worse with imputed datasets
than less accurate models, as can be expected.
The classification accuracy decreases only slightly while using imputed data in a model with low accuracy.

\section{Conclusion}
\label{sec:conclusion}
We focused on missing entire features reconstruction and its impact on the classification accuracy
of an already learned model.
We deal with traditional imputation methods: linear regression, $k$-NN, and MICE,
as well as modern methods: MLP, and XGBT.
We also proposed two methods, linear and information imputability,
for the ordering of missing features when more of them are imputed sequentially.
However, in practice information imputability is hard to estimate and does not provide satisfying results.

Comprehensive experiments are presented on four real and six artificial datasets.
The imputation influence is studied on six commonly used binary classification models: random forest,
logistic regression, $k$-NN, naive Bayes, MLP, and XGBT.
The amount of missing data varies between $10\%$ and $50\%$.

As our results indicate MICE and linear regression are generally good imputers
regardless of the amount of missingness or the classification model used.
This can be seen as some kind of generality when the used classification model is unknown.

As was also shown modern imputation methods MLP and XGBT did not perform as well as expected.
They rarely perform among the top methods. Their performance is often one of the lowest.
This result is surprising since in many current machine learning tasks these methods are one of the best.

The experimental results of this work shed more light on the applicability of
state-of-the-art imputation methods on data and their ability to reconstruct missing entire
features. The study is also important thanks to its scope of datasets, methods and portions
of missing data (up to $50\%$).


\section*{Acknowledgements}
This research was supported by SGS grant No. SGS17/210/OHK3/3T/18 and by GACR grant No. GA18-18080S.

%
%


\begin{table}[ht]
\begin{center}
\caption{Mean accuracy changes in percentages ($\pm$ standard deviation) for single missing feature imputation.
Classification methods are shown in the last six columns and imputations methods are given in the second column.}
\label{tab:single_feature}
\tiny
\begin{tabular}{|c|c|c|c|c|c|c|c|}
\hline
Dataset & Method & LR & MLP & $k$-NN & NB & XGBT & RF \\
\hline
Cancer  & knn & $ \textbf{-0.16} \pm 0.35  $ & $ 0.05  \pm 0.29  $ & $ 0.05  \pm 0.86  $ & $ 0.44  \pm 0.38  $ & $ 0.33  \pm 0.42  $ & $ \textbf{0.44} \pm 0.38  $ \\
Cancer  & linreg-iter & $ \textbf{-0.16} \pm 0.35  $ & $ 0.05  \pm 0.29  $ & $ -0.0  \pm 1.07  $ & $ 0.44  \pm 0.38  $ & $ 0.33  \pm 0.42  $ & $ \textbf{0.44} \pm 0.29  $ \\
Cancer  & linreg-li & $ \textbf{-0.16} \pm 0.35  $ & $ 0.05  \pm 0.29  $ & $ 0.00  \pm 1.07 $ & $ 0.44  \pm 0.38  $ & $ 0.33  \pm 0.45  $ & $ \textbf{0.44} \pm 0.29  $ \\
Cancer  & mice  & $ -0.22 \pm 0.61  $ & $ \textbf{0.27} \pm 0.5 $ & $ -0.22 \pm 0.89  $ & $ 0.65 \pm 0.42  $ & $ 0.11  \pm 0.41  $ & $ \textbf{0.44} \pm 0.52  $ \\
Cancer  & mlp & $ -0.38 \pm 0.73  $ & $ 0.0 \pm 0.6 $ & $ \textbf{0.11} \pm 1.03  $ & $ \textbf{0.71} \pm 0.55  $ & $ 0.16  \pm 0.49  $ & $ 0.33  \pm 0.42  $ \\
Cancer  & mlp-li  & $ -0.38 \pm 0.48  $ & $ 0.22  \pm 0.61  $ & $ \textbf{0.11} \pm 0.8 $ & $ 0.38  \pm 0.59  $ & $ -0.11 \pm 0.77  $ & $ 0.11  \pm 0.41  $ \\
Cancer  & xgb-iter  & $ \textbf{-0.11}  \pm 0.22  $ & $ 0.22  \pm 0.55  $ & $ \textbf{0.11} \pm 0.59  $ & $ 0.6 \pm 0.33  $ & $ \textbf{0.54} \pm 0.52  $ & $ 0.38  \pm 0.33  $ \\
Cancer  & xgb-li  & $ \textbf{-0.16} \pm 0.35  $ & $ 0.11  \pm 0.59  $ & $ \textbf{0.11} \pm 0.64  $ & $ 0.6 \pm 0.22  $ & $ 0.38  \pm 0.48  $ & $ 0.33  \pm 0.35  $ \\
\hline
Wine  & knn & $ -1.38 \pm 2.33  $ & $ -0.07 \pm 1.0 $ & $ -1.19 \pm 1.97  $ & $ -0.37 \pm 1.01  $ & $ -1.86 \pm 1.52  $ & $ -1.1  \pm 0.91  $ \\
Wine  & linreg-iter & $ \textbf{-0.45} \pm 0.97  $ & $ 0.2 \pm 1.01  $ & $ -0.87 \pm 2.1 $ & $ \textbf{-0.1} \pm 0.59  $ & $ -1.39 \pm 1.22  $ & $ -0.74 \pm 1.0 $ \\
Wine  & linreg-li & $ \textbf{-0.45} \pm 0.97  $ & $ 0.2 \pm 1.01  $ & $ -0.87 \pm 2.1 $ & $ \textbf{-0.1} \pm 0.59  $ & $ -1.39 \pm 1.22  $ & $ -0.74 \pm 1.0 $ \\
Wine  & mice  & $ -0.77 \pm 1.65  $ & $ -0.19 \pm 0.67  $ & $ \textbf{-0.35}  \pm 0.72  $ & $ -0.48 \pm 0.6 $ & $ \textbf{-0.79}  \pm 0.98  $ & $ \textbf{-0.54}  \pm 0.55  $ \\
Wine  & mlp & $ -3.17 \pm 5.3 $ & $ -0.28 \pm 0.73  $ & $ -1.46 \pm 2.71  $ & $ -2.23 \pm 4.94  $ & $ -3.32 \pm 3.44  $ & $ -1.84 \pm 1.76  $ \\
Wine  & mlp-li  & $ -2.2  \pm 2.98  $ & $ -0.24 \pm 0.55  $ & $ -1.78 \pm 2.98  $ & $ -2.52 \pm 6.64  $ & $ -2.7  \pm 2.55  $ & $ -1.74 \pm 1.84  $ \\
Wine  & xgb-iter  & $ -1.68 \pm 3.92  $ & $ \textbf{0.28} \pm 0.7 $ & $ -0.57 \pm 1.35  $ & $ \textbf{-1.33} \pm 3.52  $ & $ -2.43 \pm 2.97  $ & $ -1.02 \pm 2.35  $ \\
Wine  & xgb-li  & $ -1.71 \pm 4.19  $ & $ \textbf{0.22}  \pm 0.65  $ & $ -0.67 \pm 1.47  $ & $ -1.35 \pm 3.51  $ & $ -2.62 \pm 3.13  $ & $ -1.1  \pm 2.3 $ \\
\hline
MAGIC & knn & $ -0.74 \pm 1.94  $ & $ -1.18  \pm 2.58  $ & $ -1.44 \pm 2.53  $ & $ \textbf{0.01} \pm 0.54  $ & $ -2.0 \pm 2.11  $ & $ -1.84 \pm 2.49  $ \\
MAGIC & linreg-iter & $ -0.63 \pm 1.66  $ & $ \textbf{-1.12}  \pm 2.56  $ & $ \textbf{-1.27}  \pm 2.2 $ & $ -0.11 \pm 0.6 $ & $ -2.03  \pm 2.48  $ & $ -1.95 \pm 2.72  $ \\
MAGIC & linreg-li & $ -0.63 \pm 1.66  $ & $ \textbf{-1.12}  \pm 2.56  $ & $ \textbf{-1.27}  \pm 2.2 $ & $ -0.11 \pm 0.6 $ & $ -2.03  \pm 2.48  $ & $ -1.95 \pm 2.72  $ \\
MAGIC & mice  & $ -0.86 \pm 2.13  $ & $ -1.24 \pm 2.86  $ & $ -1.5  \pm 2.68  $ & $ -0.1  \pm 0.46  $ & $ \textbf{-1.88}  \pm 2.42  $ & $ \textbf{-1.36}  \pm 2.44  $ \\
MAGIC & mlp & $ -0.66 \pm 1.71  $ & $ -1.2 \pm 2.73  $ & $ -1.31  \pm 2.34  $ & $ -0.07 \pm 0.64  $ & $ -2.25 \pm 2.89  $ & $ -1.96 \pm 3.07  $ \\
MAGIC & mlp-li  & $ -0.67 \pm 1.72  $ & $ \textbf{-1.17}  \pm 2.65  $ & $ \textbf{-1.26}  \pm 2.26  $ & $ -0.11 \pm 0.59  $ & $ -2.11 \pm 2.77  $ & $ -1.83 \pm 2.83  $ \\
MAGIC & xgb-iter  & $ \textbf{-0.36}  \pm 1.31  $ & $ -1.49 \pm 2.66  $ & $ -0.95 \pm 1.66  $ & $ -0.33 \pm 0.99  $ & $ -2.1  \pm 2.65  $ & $ -2.37 \pm 3.32  $ \\
MAGIC & xgb-li  & $ \textbf{-0.38}  \pm 1.3 $ & $ -1.63 \pm 2.95  $ & $ -0.99 \pm 1.67  $ & $ -0.32 \pm 0.98  $ & $ -2.17 \pm 2.74  $ & $ -2.43 \pm 3.37  $ \\
\hline
Spambase  & knn & $ -0.15 \pm 0.64  $ & $ -0.16 \pm 0.71  $ & $ -0.36 \pm 1.69  $ & $ \textbf{-0.27} \pm 1.22  $ & $ 0.08  \pm 4.99  $ & $ -0.7  \pm 2.75  $ \\
Spambase  & linreg-iter & $ -0.11 \pm 0.66  $ & $ -0.26 \pm 1.18  $ & $ -0.34 \pm 1.76  $ & $ \textbf{-0.3}  \pm 1.02  $ & $ -0.45 \pm 1.84  $ & $ -0.24 \pm 0.79  $ \\
Spambase  & linreg-li & $ -0.11 \pm 0.66  $ & $ -0.26 \pm 1.18  $ & $ -0.34 \pm 1.76  $ & $ \textbf{-0.3}  \pm 1.02  $ & $ -0.45 \pm 1.84  $ & $ -0.24 \pm 0.79  $ \\
Spambase  & mice  & $ -0.18 \pm 0.46  $ & $ -0.18 \pm 0.4 $ & $ \textbf{-0.03}  \pm 0.35  $ & $ \textbf{-0.1} \pm 0.43  $ & $ \textbf{1.03} \pm 1.12  $ & $ \textbf{-0.2} \pm 0.78  $ \\
Spambase  & mlp & $ -0.38 \pm 0.99  $ & $ -0.49 \pm 1.65  $ & $ -0.17 \pm 0.93  $ & $ -3.6  \pm 6.18  $ & $ -0.55 \pm 2.23  $ & $ -0.36 \pm 1.28  $ \\
Spambase  & mlp-li  & $ -0.16 \pm 0.41  $ & $ -0.11 \pm 0.49  $ & $ -0.18 \pm 0.87  $ & $ -1.87 \pm 3.25  $ & $ -0.5  \pm 2.1 $ & $ -0.33 \pm 1.16  $ \\
Spambase  & xgb-iter  & $ \textbf{0.01} \pm 0.17  $ & $ \textbf{0.01} \pm 0.24  $ & $ -0.13 \pm 0.64  $ & $ \textbf{-0.09}  \pm 0.96  $ & $ -0.77 \pm 4.27  $ & $ \textbf{-0.19}  \pm 0.9 $ \\
Spambase  & xgb-li  & $ \textbf{0.01} \pm 0.21  $ & $ \textbf{0.01} \pm 0.23  $ & $ -0.08 \pm 0.46  $ & $ \textbf{-0.13} \pm 0.93  $ & $ -0.78 \pm 4.28  $ & $ -0.24 \pm 0.97  $ \\
\hline
\hline
Ringnorm  & knn & $ -0.65 \pm 0.32  $ & $ -0.47 \pm 0.47  $ & $ -1.32 \pm 0.37  $ & $ -0.5  \pm 0.17  $ & $ -0.4 \pm 0.21  $ & $ -0.16 \pm 0.19  $ \\
Ringnorm  & linreg-iter & $ -0.58 \pm 0.36  $ & $ -0.49 \pm 0.45  $ & $ -1.34 \pm 0.35  $ & $ -0.5  \pm 0.16  $ & $ -0.42 \pm 0.22  $ & $ -0.15  \pm 0.21  $ \\
Ringnorm  & linreg-li & $ -0.58 \pm 0.36  $ & $ -0.49 \pm 0.45  $ & $ -1.34 \pm 0.35  $ & $ -0.5  \pm 0.16  $ & $ -0.42  \pm 0.22  $ & $ -0.15  \pm 0.21  $ \\
Ringnorm  & mice  & $ -0.53 \pm 0.43  $ & $ \textbf{-0.14}  \pm 0.55  $ & $ \textbf{0.08} \pm 0.31  $ & $ -0.46 \pm 0.15  $ & $ -0.43 \pm 0.18  $ & $ -0.16 \pm 0.17  $ \\
Ringnorm  & mlp & $ -0.77 \pm 0.37  $ & $ -0.68 \pm 0.45  $ & $ -0.8  \pm 0.32  $ & $ \textbf{-0.41}  \pm 0.17  $ & $ \textbf{-0.39}  \pm 0.21  $ & $ \textbf{-0.14}  \pm 0.2 $ \\
Ringnorm  & mlp-li  & $ -0.75 \pm 0.45  $ & $ -0.72 \pm 0.54  $ & $ -0.74 \pm 0.33  $ & $ -0.43 \pm 0.14  $ & $ \textbf{-0.39}  \pm 0.21  $ & $ \textbf{-0.14}  \pm 0.22  $ \\
Ringnorm  & xgb-iter  & $ \textbf{-0.3} \pm 0.36  $ & $ \textbf{-0.2}  \pm 0.5 $ & $ -1.28 \pm 0.35  $ & $ -0.5  \pm 0.15  $ & $ -0.4 \pm 0.22  $ & $ -0.16 \pm 0.2 $ \\
Ringnorm  & xgb-li  & $ \textbf{-0.3} \pm 0.43  $ & $ -0.21 \pm 0.51  $ & $ -1.3  \pm 0.35  $ & $ -0.49 \pm 0.15  $ & $ \textbf{-0.38}  \pm 0.23  $ & $ -0.16 \pm 0.21  $ \\
\hline
Twonorm & knn & $ \textbf{-0.42}  \pm 0.18  $ & $ -0.43  \pm 0.21  $ & $ \textbf{-0.1} \pm 0.19  $ & $ \textbf{-0.38}  \pm 0.19  $ & $ \textbf{-0.43}  \pm 0.18  $ & $ -0.12 \pm 0.28  $ \\
Twonorm & linreg-iter & $ \textbf{-0.42}  \pm 0.21  $ & $ \textbf{-0.42}  \pm 0.18  $ & $ \textbf{-0.11}  \pm 0.18  $ & $ \textbf{-0.37}  \pm 0.2 $ & $ \textbf{-0.41}  \pm 0.19  $ & $ \textbf{-0.08}  \pm 0.33  $ \\
Twonorm & linreg-li & $ \textbf{-0.42}  \pm 0.21  $ & $ \textbf{-0.42}  \pm 0.18  $ & $ \textbf{-0.11}  \pm 0.18  $ & $ \textbf{-0.37}  \pm 0.2 $ & $ \textbf{-0.41}  \pm 0.19  $ & $ \textbf{-0.08}  \pm 0.33  $ \\
Twonorm & mice  & $ -0.58 \pm 0.21  $ & $ -0.54 \pm 0.21  $ & $ -0.21 \pm 0.18  $ & $ -0.56 \pm 0.2 $ & $ -0.55 \pm 0.21  $ & $ -0.15 \pm 0.23  $ \\
Twonorm & mlp & $ -0.45 \pm 0.23  $ & $ -0.43 \pm 0.22  $ & $ \textbf{-0.1} \pm 0.22  $ & $ -0.4 \pm 0.23  $ & $ \textbf{-0.42}  \pm 0.23  $ & $ \textbf{-0.09}  \pm 0.27  $ \\
Twonorm & mlp-li  & $ -0.47 \pm 0.19  $ & $ -0.46 \pm 0.25  $ & $ \textbf{-0.11}  \pm 0.22  $ & $ -0.45 \pm 0.19  $ & $ -0.45  \pm 0.19  $ & $ -0.14 \pm 0.24  $ \\
Twonorm & xgb-iter  & $ \textbf{-0.42}  \pm 0.19  $ & $ \textbf{-0.41}  \pm 0.19  $ & $ -0.17 \pm 0.2 $ & $ \textbf{-0.39}  \pm 0.22  $ & $ \textbf{-0.41}  \pm 0.21  $ & $ -0.33 \pm 0.25  $ \\
Twonorm & xgb-li  & $ \textbf{-0.41}  \pm 0.2 $ & $ \textbf{-0.42}  \pm 0.2 $ & $ -0.16 \pm 0.19  $ & $ \textbf{-0.39}  \pm 0.2 $ & $ \textbf{-0.43}  \pm 0.22  $ & $ -0.28 \pm 0.29  $ \\
\hline
ds\_10\_7\_3  & knn & $ \textbf{0.03}  \pm 0.13  $ & $ \textbf{-0.09} \pm 0.12  $ & $ 0.07  \pm 0.18  $ & $ -0.04 \pm 0.14  $ & $ -0.17 \pm 0.17  $ & $ -0.29 \pm 0.19  $ \\
ds\_10\_7\_3  & linreg-iter & $ \textbf{0.05} \pm 0.09  $ & $ \textbf{-0.07}  \pm 0.1 $ & $ \textbf{0.13} \pm 0.14  $ & $ -0.03 \pm 0.11  $ & $ \textbf{-0.14} \pm 0.12  $ & $ -0.29 \pm 0.18  $ \\
ds\_10\_7\_3  & linreg-li & $ \textbf{0.05}  \pm 0.09  $ & $ \textbf{-0.07}  \pm 0.1  $ & $ \textbf{0.13} \pm 0.14  $ & $ -0.03 \pm 0.11 $ & $ \textbf{-0.14} \pm 0.12  $ & $ -0.29 \pm 0.18  $ \\
ds\_10\_7\_3  & mice  & $ -0.02 \pm 0.17  $ & $ -0.19 \pm 0.23  $ & $ -0.02 \pm 0.18  $ & $ \textbf{0.07} \pm 0.13  $ & $ \textbf{-0.1} \pm 0.14  $ & $ \textbf{0.05} \pm 0.14  $ \\
ds\_10\_7\_3  & mlp & $ \textbf{0.03}  \pm 0.12  $ & $ \textbf{-0.11} \pm 0.11  $ & $ \textbf{0.11}  \pm 0.16  $ & $ -0.06 \pm 0.2 $ & $ \textbf{-0.15} \pm 0.13  $ & $ -0.21 \pm 0.12  $ \\
ds\_10\_7\_3  & mlp-li  & $ \textbf{0.04}  \pm 0.15  $ & $ \textbf{-0.08}  \pm 0.1 $ & $ 0.1 \pm 0.16  $ & $ -0.03 \pm 0.15  $ & $ -0.18 \pm 0.17  $ & $ -0.26 \pm 0.17  $ \\
ds\_10\_7\_3  & xgb-iter  & $ -0.19 \pm 0.39  $ & $ -0.56 \pm 0.6 $ & $ -0.08 \pm 0.38  $ & $ -0.19 \pm 0.67  $ & $ -0.71 \pm 0.91  $ & $ -0.55 \pm 0.53  $ \\
ds\_10\_7\_3  & xgb-li  & $ -0.15 \pm 0.21  $ & $ -0.55 \pm 0.5 $ & $ -0.11 \pm 0.38  $ & $ -0.07 \pm 0.49  $ & $ -0.62 \pm 0.58  $ & $ -0.43 \pm 0.47  $ \\
\hline
ds\_20\_14\_6 & knn & $ \textbf{-0.24}  \pm 0.21  $ & $ \textbf{-0.03} \pm 0.07  $ & $ -0.03 \pm 0.14  $ & $ -0.02 \pm 0.14  $ & $ -0.12 \pm 0.17  $ & $ \textbf{-0.06}  \pm 0.09  $ \\
ds\_20\_14\_6 & linreg-iter & $ \textbf{-0.25}  \pm 0.22  $ & $ \textbf{-0.01}  \pm 0.05  $ & $ \textbf{-0.01} \pm 0.13  $ & $ \textbf{0.03} \pm 0.12  $ & $ -0.09 \pm 0.15  $ & $ \textbf{-0.06}  \pm 0.09  $ \\
ds\_20\_14\_6 & linreg-li & $ \textbf{-0.25}  \pm 0.22 $ & $ \textbf{-0.01}  \pm 0.05  $ & $ \textbf{-0.01} \pm 0.13  $ & $ \textbf{0.03} \pm 0.12  $ & $ -0.09 \pm 0.15  $ & $ \textbf{-0.06}  \pm 0.09  $ \\
ds\_20\_14\_6 & mice  & $ -0.58 \pm 0.42  $ & $ -0.16 \pm 0.18  $ & $ -0.05 \pm 0.17  $ & $ -0.15 \pm 0.24  $ & $ \textbf{-0.01}  \pm 0.15  $ & $ \textbf{-0.06}  \pm 0.14  $ \\
ds\_20\_14\_6 & mlp & $ \textbf{-0.29} \pm 0.25  $ & $ \textbf{-0.02} \pm 0.06  $ & $ \textbf{0.01} \pm 0.12  $ & $ \textbf{0.0}  \pm 0.14  $ & $ -0.07 \pm 0.13  $ & $ \textbf{-0.07}  \pm 0.09  $ \\
ds\_20\_14\_6 & mlp-li  & $ \textbf{-0.24}  \pm 0.27  $ & $ \textbf{-0.01}  \pm 0.05  $ & $ -0.03 \pm 0.14  $ & $ -0.03 \pm 0.1 $ & $ -0.08 \pm 0.11  $ & $ \textbf{-0.07}  \pm 0.12  $ \\
ds\_20\_14\_6 & xgb-iter  & $ -0.7  \pm 0.64  $ & $ -0.21 \pm 0.25  $ & $ -0.22 \pm 0.22  $ & $ -0.23 \pm 0.37  $ & $ -0.22 \pm 0.21  $ & $ -0.21 \pm 0.27  $ \\
ds\_20\_14\_6 & xgb-li  & $ -0.76 \pm 0.7 $ & $ -0.23 \pm 0.29  $ & $ -0.18 \pm 0.24  $ & $ -0.26 \pm 0.39  $ & $ -0.27 \pm 0.3 $ & $ -0.22 \pm 0.27  $ \\
\hline
ds\_50\_35\_15  & knn & $ -0.12 \pm 0.14  $ & $ \textbf{-0.02} \pm 0.04  $ & $ -0.07 \pm 0.08  $ & $ 0.04  \pm 0.11  $ & $ \textbf{-0.02} \pm 0.05  $ & $ -0.05 \pm 0.11  $ \\
ds\_50\_35\_15  & linreg-iter & $ \textbf{-0.06}  \pm 0.09  $ & $ \textbf{-0.01}  \pm 0.02  $ & $ \textbf{-0.02}  \pm 0.04  $ & $ \textbf{0.03}  \pm 0.07  $ & $ \textbf{0.0}  \pm 0.04  $ & $ \textbf{-0.02} \pm 0.05  $ \\
ds\_50\_35\_15  & linreg-li & $ \textbf{-0.06}  \pm 0.09  $ & $ \textbf{-0.01}  \pm 0.02  $ & $ \textbf{-0.02}  \pm 0.04  $ & $ \textbf{0.03}  \pm 0.07  $ & $ \textbf{0.0}  \pm 0.04  $ & $ \textbf{-0.02} \pm 0.05  $ \\
ds\_50\_35\_15  & mice  & $ -0.09 \pm 0.15  $ & $ -0.15 \pm 0.14  $ & $ -0.26 \pm 0.22  $ & $ -0.06 \pm 0.15  $ & $ -0.03 \pm 0.15  $ & $ -0.05 \pm 0.14  $ \\
ds\_50\_35\_15  & mlp & $ -0.08 \pm 0.12  $ & $ \textbf{-0.02} \pm 0.04  $ & $ \textbf{-0.04} \pm 0.06  $ & $ \textbf{0.05} \pm 0.07  $ & $ \textbf{-0.01} \pm 0.07  $ & $ \textbf{-0.03} \pm 0.09  $ \\
ds\_50\_15\_35  & mlp-li  & $ -0.09 \pm 0.13  $ & $ \textbf{-0.01}  \pm 0.04  $ & $ -0.05 \pm 0.06  $ & $ \textbf{0.05} \pm 0.08  $ & $ \textbf{0.0}  \pm 0.06  $ & $ \textbf{-0.01}  \pm 0.08  $ \\
ds\_50\_35\_15  & xgb-iter  & $ -0.15 \pm 0.2 $ & $ -0.12 \pm 0.13  $ & $ -0.21 \pm 0.23  $ & $ -0.04 \pm 0.21  $ & $ -0.12 \pm 0.17  $ & $ -0.13 \pm 0.22  $ \\
ds\_50\_35\_15  & xgb-li  & $ -0.18 \pm 0.26  $ & $ -0.13 \pm 0.14  $ & $ -0.24 \pm 0.22  $ & $ -0.05 \pm 0.26  $ & $ -0.15 \pm 0.23  $ & $ -0.19 \pm 0.28  $ \\
\hline
ds\_100\_70\_30 & knn & $ \textbf{-0.3}  \pm 0.99  $ & $ -0.05 \pm 0.07  $ & $ -0.07 \pm 0.12  $ & $ -0.04 \pm 0.14  $ & $ \textbf{0.03} \pm 0.16  $ & $ \textbf{-0.02} \pm 0.08  $ \\
ds\_100\_70\_30 & linreg-iter & $ \textbf{0.0} \pm 0.14  $ & $ \textbf{0.0}  \pm 0.01  $ & $ \textbf{-0.02}  \pm 0.06  $ & $ 0.0 \pm 0.04  $ & $ \textbf{0.01}  \pm 0.05  $ & $ \textbf{-0.01}  \pm 0.06  $ \\
ds\_100\_70\_30 & linreg-li & $ \textbf{0.0} \pm 0.14  $ & $ \textbf{0.0}  \pm 0.01  $ & $ \textbf{-0.02}  \pm 0.06  $ & $ 0.0 \pm 0.04  $ & $ \textbf{0.01}  \pm 0.05  $ & $ \textbf{-0.01}  \pm 0.06  $ \\
ds\_100\_70\_30 & mice  & $ -1.76 \pm 1.85  $ & $ -0.25 \pm 0.18  $ & $ -0.09 \pm 0.17  $ & $ \textbf{0.02}  \pm 0.18  $ & $ -0.14 \pm 0.22  $ & $ \textbf{-0.02} \pm 0.12  $ \\
ds\_100\_70\_30 & mlp & $ \textbf{-0.03} \pm 0.35  $ & $ -0.05 \pm 0.07  $ & $ -0.05 \pm 0.09  $ & $ 0.0 \pm 0.09  $ & $ \textbf{0.01}  \pm 0.08  $ & $ \textbf{-0.01}  \pm 0.1 $ \\
ds\_100\_70\_30 & mlp-li  & $ \textbf{0.03} \pm 0.32  $ & $ -0.05 \pm 0.07  $ & $ -0.05 \pm 0.1 $ & $ \textbf{0.01} \pm 0.08  $ & $ -0.01 \pm 0.1 $ & $ \textbf{-0.01}  \pm 0.06  $ \\
ds\_100\_70\_30 & xgb-iter  & $ -3.67 \pm 3.86  $ & $ -0.22 \pm 0.18  $ & $ -0.13 \pm 0.29  $ & $ -0.05 \pm 0.23  $ & $ -0.16 \pm 0.31  $ & $ -0.08 \pm 0.18  $ \\
ds\_100\_70\_30 & xgb-li  & $ -3.58 \pm 4.25  $ & $ -0.22 \pm 0.18  $ & $ -0.15 \pm 0.27  $ & $ -0.06 \pm 0.22  $ & $ -0.15 \pm 0.32  $ & $ -0.08 \pm 0.19  $ \\
\hline
\end{tabular}
\end{center}
\end{table}

\begin{table}[t]
\begin{center}
\caption{Mean accuracy changes shown in percentages ($\pm$ standard deviation) on real datasets with missingness from
$10\%$ up to $50\%$ where only the two best classification models for each dataset are shown.}
\label{tab:multiple_real}
\tiny
    \begin{tabular}{| c | c | c | c | c | c | c | c |}
\hline
Dataset & Model & Method & 10\% & 20\% & 30\% & 40\% & 50\% \\
\hline
Cancer  & LR  & knn & $ -0.16 \pm 0.35  $ & $ -0.37 \pm 0.31  $ & $ -0.49 \pm 0.45  $ & $ -0.66 \pm 0.7 $ & $ -0.98 \pm 0.95  $ \\
Cancer  & LR  & linreg-iter & $ -0.16 \pm 0.35  $ & $ -0.29 \pm 0.4 $ & $ -0.43 \pm 0.56  $ & $ -0.68 \pm 0.7 $ & $ -1.06 \pm 1.01  $ \\
Cancer  & LR  & linreg-li & $ -0.16 \pm 0.35  $ & $ \textbf{-0.27} \pm 0.34  $ & $ \textbf{-0.37}  \pm 0.47  $ & $ -0.47 \pm 0.72  $ & $ -1.1  \pm 0.98  $ \\
Cancer  & LR  & mice  & $ -0.22 \pm 0.61  $ & $ \textbf{-0.27}  \pm 0.63  $ & $ -0.61 \pm 0.76  $ & $ -0.98 \pm 0.99  $ & $ -1.32 \pm 1.1 $ \\
Cancer  & LR  & mlp & $ -0.38 \pm 0.73  $ & $ -0.44 \pm 0.61  $ & $ -0.63 \pm 0.69  $ & $ -0.81 \pm 1.03  $ & $ -1.14 \pm 1.06  $ \\
Cancer  & LR  & mlp-li  & $ -0.38 \pm 0.48  $ & $ -0.59 \pm 0.49  $ & $ -0.74 \pm 0.74  $ & $ -1.05 \pm 0.82  $ & $ -1.54 \pm 1.74  $ \\
Cancer  & LR  & xgb-iter  & $ \textbf{-0.11}  \pm 0.22  $ & $ -0.32 \pm 0.37  $ & $ -0.54 \pm 0.5 $ & $ -0.47 \pm 0.61  $ & $ -1.0  \pm 1.15  $ \\
Cancer  & LR  & xgb-li  & $ -0.16 \pm 0.35  $ & $ \textbf{-0.25}  \pm 0.46  $ & $ -0.56 \pm 0.83  $ & $ \textbf{-0.29}  \pm 0.64  $ & $ \textbf{-0.78} \pm 0.99  $ \\
\hline
Cancer  & RF  & knn & $ \textbf{0.44} \pm 0.38  $ & $ \textbf{0.49} \pm 0.45  $ & $ 0.22  \pm 0.77  $ & $ 0.12  \pm 0.91  $ & $ -0.69 \pm 1.52  $ \\
Cancer  & RF  & linreg-iter & $ \textbf{0.44} \pm 0.29  $ & $ \textbf{0.5}  \pm 0.41  $ & $ \textbf{0.33}  \pm 0.66  $ & $ -0.03 \pm 0.85  $ & $ \textbf{-0.63} \pm 1.24  $ \\
Cancer  & RF  & linreg-li & $ \textbf{0.44} \pm 0.29  $ & $ 0.39  \pm 0.41  $ & $ \textbf{0.42} \pm 0.75 $ & $ \textbf{0.17} \pm 0.8  $ & $ \textbf{-0.66}  \pm 1.14  $ \\
Cancer  & RF  & mice  & $ \textbf{0.44} \pm 0.52  $ & $ 0.37  \pm 0.67  $ & $ -0.2  \pm 0.86  $ & $ -0.83 \pm 1.35  $ & $ -1.23 \pm 0.95  $ \\
Cancer  & RF  & mlp & $ 0.33  \pm 0.42  $ & $ 0.1 \pm 0.8 $ & $ 0.02  \pm 0.8 $ & $ -0.46 \pm 1.0 $ & $ -1.08 \pm 1.35  $ \\
Cancer  & RF  & mlp-li  & $ 0.11  \pm 0.41  $ & $ -0.12 \pm 0.71  $ & $ -0.47 \pm 1.14  $ & $ -1.3  \pm 1.57  $ & $ -2.72 \pm 4.01  $ \\
Cancer  & RF  & xgb-iter  & $ 0.38  \pm 0.33  $ & $ 0.29  \pm 0.56  $ & $ 0.12  \pm 0.61  $ & $ -0.12 \pm 0.79  $ & $ \textbf{-0.44} \pm 1.1 $ \\
Cancer  & RF  & xgb-li  & $ 0.33  \pm 0.35  $ & $ 0.42  \pm 0.43  $ & $ 0.22  \pm 0.83  $ & $ 0.1 \pm 0.82  $ & $ \textbf{-0.66} \pm 0.7 $ \\
\hline
Wine  & XGBT  & knn & $ -1.86 \pm 1.52  $ & $ -3.21 \pm 1.25  $ & $ -5.57 \pm 2.36  $ & $ -7.42 \pm 2.4 $ & $ -9.23 \pm 2.81  $ \\
Wine  & XGBT  & linreg-iter & $ -1.39 \pm 1.22 $ & $ -2.9  \pm 1.53  $ & $ -3.36 \pm 1.66  $ & $ -6.78 \pm 2.44  $ & $ -7.61 \pm 2.59  $ \\
Wine  & XGBT  & linreg-li & $ -1.39 \pm 1.22  $ & $ -2.94 \pm 1.29  $ & $ -4.54 \pm 2.5  $ & $ -5.95 \pm 2.4  $ & $ -7.5 \pm 3.08  $ \\
Wine  & XGBT  & mice  & $ \textbf{-0.79}  \pm 0.98  $ & $ \textbf{-1.15}  \pm 0.97  $ & $ \textbf{-2.45}  \pm 1.95  $ & $ \textbf{-3.75}  \pm 2.25  $ & $ \textbf{-4.13}  \pm 2.67  $ \\
Wine  & XGBT  & mlp & $ -3.42 \pm 2.67  $ & $ -5.49 \pm 3.86  $ & $ -7.69 \pm 4.55  $ & $ -10.75  \pm 5.12  $ & $ -13.25  \pm 5.57  $ \\
Wine  & XGBT  & mlp-li  & $ -3.51 \pm 3.78  $ & $ -5.91 \pm 3.88  $ & $ -8.4  \pm 4.1 $ & $ -10.33  \pm 4.37  $ & $ -12.29  \pm 4.97  $ \\
Wine  & XGBT  & xgb-iter  & $ -2.43 \pm 2.97  $ & $ -5.55 \pm 4.53  $ & $ -6.72 \pm 5.75  $ & $ -9.57 \pm 5.51  $ & $ -14.86  \pm 6.54  $ \\
Wine  & XGBT  & xgb-li  & $ -2.62 \pm 3.13  $ & $ -5.07 \pm 4.38  $ & $ -8.17 \pm 4.82  $ & $ -9.75 \pm 4.31  $ & $ -11.57  \pm 5.83  $ \\
\hline
Wine  & RF  & knn & $ -1.1  \pm 0.91  $ & $ -1.94 \pm 0.78  $ & $ -3.61 \pm 1.44  $ & $ -4.72 \pm 1.63  $ & $ -5.8  \pm 1.73  $ \\
Wine  & RF  & linreg-iter & $ -0.74 \pm 1.0  $ & $ -1.5  \pm 1.19  $ & $ \textbf{-1.73}  \pm 1.12  $ & $ -3.88 \pm 1.8 $ & $ -5.29 \pm 1.83  $ \\
Wine  & RF  & linreg-li & $ -0.74 \pm 1.0  $ & $ -1.52  \pm 1.2  $ & $ -2.7  \pm 2.0  $ & $ -3.84 \pm 1.73 $ & $ -5.49 \pm 2.6  $ \\
Wine  & RF  & mice  & $ \textbf{-0.54}  \pm 0.55  $ & $ \textbf{-0.95}  \pm 0.61  $ & $ \textbf{-1.79}  \pm 0.99  $ & $ \textbf{-2.64}  \pm 1.59  $ & $ \textbf{-2.87}  \pm 1.71  $ \\
Wine  & RF  & mlp & $ -1.69 \pm 1.65  $ & $ -2.95 \pm 2.23  $ & $ -5.26 \pm 2.47  $ & $ -7.24 \pm 2.05  $ & $ -8.7  \pm 1.82  $ \\
Wine  & RF  & mlp-li  & $ -1.74 \pm 1.5 $ & $ -3.32 \pm 1.86  $ & $ -5.12 \pm 2.03  $ & $ -6.12 \pm 2.21  $ & $ -8.44 \pm 1.76  $ \\
Wine  & RF  & xgb-iter  & $ -1.02 \pm 2.35  $ & $ -3.3  \pm 3.05  $ & $ -4.5  \pm 3.26  $ & $ -6.14 \pm 2.53  $ & $ -8.0  \pm 2.94  $ \\
Wine  & RF  & xgb-li  & $ -1.1  \pm 2.3 $ & $ -2.65 \pm 3.01  $ & $ -5.33 \pm 3.39  $ & $ -6.22 \pm 2.61  $ & $ -7.58 \pm 3.26  $ \\
\hline
MAGIC & XGBT  & knn & $ \textbf{-2.0} \pm 2.11  $ & $ -3.05 \pm 2.26  $ & $ \textbf{-5.3}  \pm 3.07  $ & $ -8.06 \pm 4.47  $ & $ -10.71  \pm 4.19  $ \\
MAGIC & XGBT  & linreg-iter & $ \textbf{-2.03}  \pm 2.48  $ & $ -4.19 \pm 3.52  $ & $ -6.28 \pm 4.82  $ & $ -9.99 \pm 6.0 $ & $ -12.82  \pm 5.76  $ \\
MAGIC & XGBT  & linreg-li & $ \textbf{-2.03}  \pm 2.48  $ & $ -4.68 \pm 3.86  $ & $ -6.77 \pm 4.91 $ & $ -9.2 \pm 6.2  $ & $ -11.25 \pm 4.96  $ \\
MAGIC & XGBT  & mice  & $ \textbf{-2.03}  \pm 2.52  $ & $ \textbf{-2.51}  \pm 2.25  $ & $ \textbf{-4.78}  \pm 3.39  $ & $ \textbf{-7.02}  \pm 4.09  $ & $ \textbf{-8.52}  \pm 3.55  $ \\
MAGIC & XGBT  & mlp & $ -5.46 \pm 7.85  $ & $ -3.94 \pm 2.96  $ & $ \textbf{-4.99}  \pm 3.39  $ & $ -9.59 \pm 5.69  $ & $ -11.77  \pm 7.9 $ \\
MAGIC & XGBT  & mlp-li  & $ -2.26 \pm 2.9 $ & $ -4.25 \pm 3.59  $ & $ -6.95 \pm 4.43  $ & $ \textbf{-7.77} \pm 4.76  $ & $ \textbf{-8.59}  \pm 4.55  $ \\
MAGIC & XGBT  & xgb-iter  & $ -3.99 \pm 6.05  $ & $ -4.94 \pm 6.29  $ & $ -10.53  \pm 7.01  $ & $ -14.97  \pm 7.45  $ & $ -15.49  \pm 9.18  $ \\
MAGIC & XGBT  & xgb-li  & $ \textbf{-2.34} \pm 2.85  $ & $ -3.98 \pm 3.18  $ & $ -9.45 \pm 6.46  $ & $ -9.13 \pm 9.71  $ & $ -16.31  \pm 13.06 $ \\
\hline
MAGIC & RF  & knn & $ -2.7  \pm 2.49  $ & $ -3.76 \pm 2.54  $ & $ -5.79 \pm 3.19  $ & $ -8.37 \pm 4.33  $ & $ -10.7 \pm 3.81  $ \\
MAGIC & RF  & linreg-iter & $ \textbf{-1.95} \pm 2.72  $ & $ -4.77 \pm 3.4 $ & $ -6.65 \pm 4.32  $ & $ -10.73  \pm 5.42  $ & $ -12.84  \pm 4.58  $ \\
MAGIC & RF  & linreg-li & $ \textbf{-1.95} \pm 2.72  $ & $ -5.38 \pm 4.0  $ & $ -7.64 \pm 4.75  $ & $ -9.47 \pm 5.73  $ & $ -11.53  \pm 4.25  $ \\
MAGIC & RF  & mice  & $ -2.36 \pm 2.54 $ & $ \textbf{-2.73} \pm 2.19 $ & $ \textbf{-4.82} \pm 3.27 $ & $ \textbf{-6.84} \pm 4.05  $ & $ \textbf{-8.42} \pm 3.55 $ \\
MAGIC & RF  & mlp & $ -5.85 \pm 7.19  $ & $ -4.41 \pm 3.1 $ & $ \textbf{-5.23}  \pm 3.19  $ & $ -9.62 \pm 5.14  $ & $ -11.25  \pm 6.48  $ \\
MAGIC & RF  & mlp-li  & $ -2.89 \pm 2.92  $ & $ -4.71 \pm 3.7 $ & $ -7.55 \pm 4.59  $ & $ -8.23 \pm 4.8 $ & $ \textbf{-9.08}  \pm 4.46  $ \\
MAGIC & RF  & xgb-iter  & $ -5.06 \pm 5.88  $ & $ -5.8  \pm 6.55  $ & $ -11.75  \pm 6.02  $ & $ -16.19  \pm 6.39  $ & $ -16.99  \pm 8.59  $ \\
MAGIC & RF  & xgb-li  & $ -3.57 \pm 3.44  $ & $ -5.61 \pm 3.94  $ & $ -10.93  \pm 6.34  $ & $ -9.92 \pm 8.83  $ & $ -17.16  \pm 11.48 $ \\
\hline
Spambase  & MLP & knn & $ -1.55 \pm 1.75  $ & $ -3.72 \pm 3.34  $ & $ -7.87 \pm 4.55  $ & $ -10.01  \pm 4.39  $ & $ -13.55  \pm 6.36  $ \\
Spambase  & MLP & linreg-iter & $ -2.24 \pm 2.56  $ & $ -3.36 \pm 2.43  $ & $ -4.91 \pm 2.72  $ & $ -6.25 \pm 2.61  $ & $ -8.34 \pm 3.12  $ \\
Spambase  & MLP & linreg-li & $ -2.24 \pm 2.56  $ & $ -2.89  \pm 2.39  $ & $ -5.36 \pm 2.7  $ & $ -8.22 \pm 4.09  $ & $ -8.76 \pm 3.28  $ \\
Spambase  & MLP & mice  & $ -1.34 \pm 1.24  $ & $ -2.33 \pm 1.39  $ & $ -4.67 \pm 1.78  $ & $ \textbf{-4.96}  \pm 2.36  $ & $ \textbf{-6.56}  \pm 2.31  $ \\
Spambase  & MLP & mlp & $ -3.44 \pm 5.06  $ & $ -7.17 \pm 5.49  $ & $ -9.64 \pm 6.78  $ & $ -11.38  \pm 7.33  $ & $ -13.9 \pm 8.77  $ \\
Spambase  & MLP & mlp-li  & $ -3.54 \pm 3.47  $ & $ -6.04 \pm 6.09  $ & $ -7.71 \pm 4.71  $ & $ -10.96  \pm 5.02  $ & $ -13.75  \pm 7.69  $ \\
Spambase  & MLP & xgb-iter  & $ \textbf{-0.65}  \pm 0.62  $ & $ -1.95 \pm 1.31  $ & $ -6.61 \pm 6.25  $ & $ -7.33 \pm 5.25  $ & $ -11.99  \pm 7.68  $ \\
Spambase  & MLP & xgb-li  & $ -0.89 \pm 0.87  $ & $ \textbf{-1.54}  \pm 1.57  $ & $ \textbf{-4.06}  \pm 3.57  $ & $ -6.41 \pm 4.59  $ & $ -7.5  \pm 3.02  $ \\
\hline
Spambase  & RF  & knn & $ -5.43 \pm 6.99  $ & $ -10.52  \pm 9.7 $ & $ -15.72  \pm 8.41  $ & $ -19.97  \pm 8.63  $ & $ -22.17  \pm 9.06  $ \\
Spambase  & RF  & linreg-iter & $ -2.05 \pm 2.29  $ & $ \textbf{-3.07} \pm 2.57  $ & $ \textbf{-4.97}  \pm 3.51  $ & $ -7.46 \pm 4.47  $ & $ -11.45  \pm 3.65  $ \\
Spambase  & RF  & linreg-li & $ -2.05 \pm 2.29  $ & $ -3.57  \pm 3.32  $ & $ \textbf{-5.01} \pm 2.55  $ & $ -8.91  \pm 5.39  $ & $ -9.09 \pm 3.73  $ \\
Spambase  & RF  & mice  & $ \textbf{-1.64} \pm 2.15  $ & $ \textbf{-2.52}  \pm 2.98  $ & $ \textbf{-5.05}  \pm 2.54  $ & $ \textbf{-4.95}  \pm 3.19  $ & $ \textbf{-5.96}  \pm 3.0 $ \\
Spambase  & RF  & mlp & $ -4.95 \pm 5.32  $ & $ -7.27 \pm 5.49  $ & $ -12.03  \pm 8.2 $ & $ -13.83  \pm 6.76  $ & $ -18.68  \pm 8.24  $ \\
Spambase  & RF  & mlp-li  & $ -3.58 \pm 3.87  $ & $ -6.24 \pm 6.33  $ & $ -10.54  \pm 7.14  $ & $ -13.37  \pm 8.97  $ & $ -18.26  \pm 8.02  $ \\
Spambase  & RF  & xgb-iter  & $ \textbf{-1.37}  \pm 2.08  $ & $ -4.54 \pm 7.02  $ & $ -10.54  \pm 11.71 $ & $ -17.62  \pm 10.25 $ & $ -21.63  \pm 9.67  $ \\
Spambase  & RF  & xgb-li  & $ -3.71 \pm 5.42  $ & $ -3.89 \pm 5.3 $ & $ -8.38 \pm 8.34  $ & $ -14.68  \pm 11.33 $ & $ -14.78  \pm 6.69  $ \\
\hline
\end{tabular}
\end{center}
\end{table}


\begin{table}[t]
\begin{center}
\caption{Mean accuracy changes shown in percentages ($\pm$ standard deviation) on artificial datasets with missingness from
$10\%$ up to $50\%$ where only the two best classification models for each dataset are shown.}
\label{tab:multiple_artificial}
\tiny
    \begin{tabular}{| c | c | c | c | c | c | c | c |}
\hline
Dataset & Model & Method & 10\% & 20\% & 30\% & 40\% & 50\% \\
\hline


Ringnorm  & NB  & knn & $ -1.16 \pm 0.25  $ & $ -3.56 \pm 0.3 $ & $ -7.31 \pm 0.53  $ & $ -13.01  \pm 0.58  $ & $ -20.59  \pm 0.65  $ \\
Ringnorm  & NB  & linreg-iter & $ -1.81 \pm 2.66  $ & $ -3.51 \pm 0.36  $ & $ -7.41 \pm 0.36  $ & $ -13.25  \pm 0.42  $ & $ -20.92  \pm 0.49  $ \\
Ringnorm  & NB  & linreg-li & $ -1.23 \pm 0.27  $ & $ -3.46 \pm 0.39 $ & $ -7.75  \pm 1.45  $ & $ -13.0  \pm 1.55  $ & $ -20.96  \pm 0.45  $ \\
Ringnorm  & NB  & mice  & $ \textbf{-1.03}  \pm 0.24  $ & $ -3.04  \pm 0.25  $ & $ \textbf{-5.8} \pm 0.45  $ & $ \textbf{-9.3} \pm 0.4 $ & $ \textbf{-16.98} \pm 3.92  $ \\
Ringnorm  & NB  & mlp & $ -1.16 \pm 0.16  $ & $ -3.34 \pm 0.94  $ & $ -6.83 \pm 0.46  $ & $ -12.51  \pm 1.67  $ & $ -19.58  \pm 0.61  $ \\
Ringnorm  & NB  & mlp-li  & $ -1.87 \pm 3.69  $ & $ \textbf{-2.95}  \pm 0.33  $ & $ -6.2  \pm 1.34  $ & $ -11.0 \pm 0.4 $ & $ -18.01 \pm 0.68  $ \\
Ringnorm  & NB  & xgb-iter  & $ -1.24 \pm 0.27  $ & $ -3.58 \pm 0.41  $ & $ -7.51 \pm 0.37  $ & $ -12.68  \pm 2.7 $ & $ -19.43  \pm 4.62  $ \\
Ringnorm  & NB  & xgb-li  & $ -1.21 \pm 0.26  $ & $ -3.42 \pm 0.19  $ & $ -7.18 \pm 0.52  $ & $ -13.05  \pm 0.45  $ & $ -20.54  \pm 0.53  $ \\
\hline
Ringnorm  & XGBT  & knn & $ -1.0  \pm 0.29  $ & $ -2.38  \pm 0.37  $ & $ -4.91 \pm 0.62  $ & $ -8.32 \pm 0.82  $ & $ -12.05  \pm 1.15  $ \\
Ringnorm  & XGBT  & linreg-iter & $ -1.34 \pm 1.5 $ & $ -2.54 \pm 0.31  $ & $ -4.91 \pm 0.52  $ & $ -8.12 \pm 0.91  $ & $ -12.42  \pm 1.46  $ \\
Ringnorm  & XGBT  & linreg-li & $ -0.95 \pm 0.28  $ & $ -2.62  \pm 0.36  $ & $ -5.37 \pm 0.88 $ & $ -8.22 \pm 1.44  $ & $ -12.37  \pm 1.05  $ \\
Ringnorm  & XGBT  & mice  & $ -0.96 \pm 0.29  $ & $ \textbf{-2.3} \pm 0.33  $ & $ \textbf{-4.32}  \pm 0.53  $ & $ \textbf{-6.81}  \pm 0.59  $ & $ \textbf{-10.76} \pm 1.9 $ \\
Ringnorm  & XGBT  & mlp & $ \textbf{-0.92}  \pm 0.27  $ & $ -2.48  \pm 0.58  $ & $ -4.57  \pm 0.53  $ & $ -7.71 \pm 1.28  $ & $ -11.4  \pm 1.14  $ \\
Ringnorm  & XGBT  & mlp-li  & $ -1.34 \pm 1.97  $ & $ -2.35  \pm 0.41  $ & $ -4.75 \pm 0.9 $ & $ -7.37  \pm 0.83  $ & $ -10.96 \pm 1.1 $ \\
Ringnorm  & XGBT  & xgb-iter  & $ \textbf{-0.89}  \pm 0.23  $ & $ \textbf{-2.29}  \pm 0.38  $ & $ -4.76 \pm 0.51  $ & $ -7.67 \pm 1.77  $ & $ -11.12 \pm 2.74  $ \\
Ringnorm  & XGBT  & xgb-li  & $ \textbf{-0.89}  \pm 0.28  $ & $ -2.38  \pm 0.36  $ & $ -4.85 \pm 0.62  $ & $ -7.88 \pm 0.76  $ & $ -12.28  \pm 1.38  $ \\
\hline
Twonorm & LR  & knn & $ -0.85 \pm 0.28  $ & $ -1.69 \pm 0.35  $ & $ -2.88 \pm 0.39  $ & $ -4.13 \pm 0.35  $ & $ \textbf{-5.94} \pm 0.58  $ \\
Twonorm & LR  & linreg-iter & $ \textbf{-0.71}  \pm 0.27  $ & $ \textbf{-1.43}  \pm 0.28  $ & $ \textbf{-2.57} \pm 0.27  $ & $ \textbf{-4.08} \pm 0.41  $ & $ \textbf{-5.84} \pm 0.57  $ \\
Twonorm & LR  & linreg-li & $ -0.78  \pm 0.3  $ & $ -1.69  \pm 0.31  $ & $ \textbf{-2.61}  \pm 0.47  $ & $ \textbf{-3.97}  \pm 0.28  $ & $ -6.03  \pm 0.48  $ \\
Twonorm & LR  & mice  & $ -1.09 \pm 0.27  $ & $ -2.19 \pm 0.4 $ & $ -3.75 \pm 0.46  $ & $ -5.27 \pm 0.43  $ & $ -7.53 \pm 0.65  $ \\
Twonorm & LR  & mlp & $ -0.91 \pm 0.15  $ & $ -1.79 \pm 0.26  $ & $ -2.7  \pm 0.27  $ & $ \textbf{-4.05} \pm 0.56  $ & $ \textbf{-5.93} \pm 0.52  $ \\
Twonorm & LR  & mlp-li  & $ -0.86 \pm 0.25  $ & $ -1.73 \pm 0.26  $ & $ -2.8  \pm 0.34  $ & $ -4.27 \pm 0.42  $ & $ -6.22 \pm 0.53  $ \\
Twonorm & LR  & xgb-iter  & $ -0.78  \pm 0.19  $ & $ -1.66 \pm 0.32  $ & $ -3.0  \pm 0.33  $ & $ -4.4  \pm 0.4 $ & $ -6.92 \pm 0.52  $ \\
Twonorm & LR  & xgb-li  & $ -0.93 \pm 0.24  $ & $ -1.74 \pm 0.31  $ & $ -2.75 \pm 0.36  $ & $ -4.25 \pm 0.36  $ & $ -6.49 \pm 0.56  $ \\
\hline
Twonorm & XGBT  & knn & $ -0.82 \pm 0.29  $ & $ -1.68 \pm 0.37  $ & $ -2.85 \pm 0.36  $ & $ -4.15 \pm 0.36  $ & $ \textbf{-5.91} \pm 0.6 $ \\
Twonorm & XGBT  & linreg-iter & $ \textbf{-0.73}  \pm 0.25  $ & $ \textbf{-1.45}  \pm 0.28  $ & $ \textbf{-2.58} \pm 0.26  $ & $ \textbf{-4.05} \pm 0.42  $ & $ \textbf{-5.82} \pm 0.56  $ \\
Twonorm & XGBT  & linreg-li & $ \textbf{-0.75}  \pm 0.32  $ & $ -1.65  \pm 0.31  $ & $ \textbf{-2.6}  \pm 0.44  $ & $ \textbf{-3.96}  \pm 0.29  $ & $ -6.02  \pm 0.49  $ \\
Twonorm & XGBT  & mice  & $ -1.1  \pm 0.27  $ & $ -2.2  \pm 0.37  $ & $ -3.73 \pm 0.46  $ & $ -5.22 \pm 0.45  $ & $ -7.56 \pm 0.65  $ \\
Twonorm & XGBT  & mlp & $ -0.87 \pm 0.17  $ & $ -1.72 \pm 0.26  $ & $ -2.71 \pm 0.23  $ & $ \textbf{-4.04} \pm 0.54  $ & $ \textbf{-5.95} \pm 0.52  $ \\
Twonorm & XGBT  & mlp-li  & $ -0.82 \pm 0.21  $ & $ -1.73 \pm 0.26  $ & $ -2.78 \pm 0.34  $ & $ -4.27 \pm 0.42  $ & $ -6.21 \pm 0.53  $ \\
Twonorm & XGBT  & xgb-iter  & $ -0.8 \pm 0.19  $ & $ -1.66 \pm 0.31  $ & $ -2.98 \pm 0.34  $ & $ -4.36 \pm 0.4 $ & $ -6.95 \pm 0.51  $ \\
Twonorm & XGBT  & xgb-li  & $ -0.92 \pm 0.2 $ & $ -1.71 \pm 0.3 $ & $ -2.73 \pm 0.37  $ & $ -4.25 \pm 0.34  $ & $ -6.53 \pm 0.57  $ \\
\hline
ds\_10\_7\_3  & MLP & knn & $ \textbf{0.25} \pm 0.12  $ & $ \textbf{0.06}  \pm 0.26  $ & $ -1.63 \pm 1.78  $ & $ -9.15 \pm 3.16  $ & $ -13.8 \pm 2.29  $ \\
ds\_10\_7\_3  & MLP & linreg-iter & $ -0.07 \pm 0.1 $ & $ \textbf{-0.03} \pm 0.53  $ & $ \textbf{-0.47} \pm 1.02  $ & $ -5.83 \pm 2.68  $ & $ -11.12  \pm 3.84  $ \\
ds\_10\_7\_3  & MLP & linreg-li & $ -0.07 \pm 0.1  $ & $ \textbf{0.01} \pm 0.5 $ & $ -1.3  \pm 2.39  $ & $ -6.02  \pm 2.48  $ & $ -11.16  \pm 4.06  $ \\
ds\_10\_7\_3  & MLP & mice  & $ 0.15  \pm 0.23  $ & $ \textbf{-0.05} \pm 0.29  $ & $ \textbf{-0.4}  \pm 0.45  $ & $ \textbf{-1.55} \pm 0.69  $ & $ \textbf{-2.51} \pm 0.71  $ \\
ds\_10\_7\_3  & MLP & mlp & $ \textbf{0.22}  \pm 0.11  $ & $ \textbf{0.1} \pm 0.23  $ & $ -0.83 \pm 1.03  $ & $ -3.4  \pm 1.54  $ & $ -7.69 \pm 2.71  $ \\
ds\_10\_7\_3  & MLP & mlp-li  & $ \textbf{0.25} \pm 0.1 $ & $ \textbf{0.01}  \pm 0.23  $ & $ -0.9  \pm 1.47  $ & $ -4.71 \pm 2.12  $ & $ -8.48 \pm 3.68  $ \\
ds\_10\_7\_3  & MLP & xgb-iter  & $ -0.23 \pm 0.6 $ & $ -1.41 \pm 1.46  $ & $ -2.28 \pm 1.69  $ & $ -4.73 \pm 2.24  $ & $ -9.07 \pm 2.55  $ \\
ds\_10\_7\_3  & MLP & xgb-li  & $ -0.22 \pm 0.5 $ & $ -0.91 \pm 0.67  $ & $ -2.29 \pm 1.8 $ & $ -3.88 \pm 1.86  $ & $ -7.14 \pm 2.71  $ \\
\hline
ds\_10\_7\_3  & XGBT  & knn & $ -0.37 \pm 0.17  $ & $ -0.67 \pm 0.31  $ & $ -2.78 \pm 2.11  $ & $ -8.77 \pm 4.0 $ & $ -13.14  \pm 3.26  $ \\
ds\_10\_7\_3  & XGBT  & linreg-iter & $ \textbf{-0.14} \pm 0.12  $ & $ -0.78 \pm 0.5 $ & $ -1.24 \pm 0.78  $ & $ -6.01 \pm 2.71  $ & $ -10.82  \pm 3.51  $ \\
ds\_10\_7\_3  & XGBT  & linreg-li & $ \textbf{-0.14} \pm 0.12  $ & $ -0.72  \pm 0.52  $ & $ -1.87  \pm 2.12 $ & $ -6.53  \pm 2.5  $ & $ -11.45  \pm 4.13  $ \\
ds\_10\_7\_3  & XGBT  & mice  & $ -0.3 \pm 0.14  $ & $ \textbf{-0.49}  \pm 0.28  $ & $ \textbf{-0.83}  \pm 0.54  $ & $ \textbf{-1.53} \pm 0.86  $ & $ \textbf{-2.61}  \pm 0.65  $ \\
ds\_10\_7\_3  & XGBT  & mlp & $ -0.35 \pm 0.13  $ & $ \textbf{-0.61} \pm 0.28  $ & $ -1.61 \pm 0.94  $ & $ -4.0  \pm 1.74  $ & $ -8.31 \pm 2.92  $ \\
ds\_10\_7\_3  & XGBT  & mlp-li  & $ -0.38 \pm 0.17  $ & $ -0.71 \pm 0.31  $ & $ -1.81 \pm 1.51  $ & $ -5.03 \pm 2.21  $ & $ -9.2  \pm 4.22  $ \\
ds\_10\_7\_3  & XGBT  & xgb-iter  & $ -0.91 \pm 0.91  $ & $ -2.2  \pm 1.26  $ & $ -3.22 \pm 2.01  $ & $ -5.16 \pm 2.48  $ & $ -10.34  \pm 2.8 $ \\
ds\_10\_7\_3  & XGBT  & xgb-li  & $ -0.82 \pm 0.58  $ & $ -1.67 \pm 1.16  $ & $ -3.19 \pm 1.74  $ & $ -5.19 \pm 1.95  $ & $ -8.28 \pm 2.5 $ \\
\hline
ds\_20\_14\_6 & MLP & knn & $ 0.02  \pm 0.26  $ & $ -0.42 \pm 0.33  $ & $ -0.71 \pm 0.39  $ & $ \textbf{-1.53} \pm 0.51  $ & $ \textbf{-2.16} \pm 0.69  $ \\
ds\_20\_14\_6 & MLP & linreg-iter & $ \textbf{0.24} \pm 0.07  $ & $ \textbf{0.18}  \pm 0.1 $ & $ \textbf{-0.19} \pm 0.52  $ & $ -4.21 \pm 1.46  $ & $ -10.0 \pm 1.75  $ \\
ds\_20\_14\_6 & MLP & linreg-li & $ \textbf{0.24} \pm 0.05  $ & $ \textbf{0.16} \pm 0.07  $ & $ \textbf{-0.18} \pm 0.87  $ & $ -3.83 \pm 1.23  $ & $ -9.83  \pm 3.27  $ \\
ds\_20\_14\_6 & MLP & mice  & $ 0.05  \pm 0.17  $ & $ -0.6  \pm 0.63  $ & $ -2.72 \pm 1.24  $ & $ -5.71 \pm 1.09  $ & $ -9.98 \pm 2.2 $ \\
ds\_20\_14\_6 & MLP & mlp & $ \textbf{0.24} \pm 0.08  $ & $ \textbf{0.11}  \pm 0.28  $ & $ \textbf{-0.15} \pm 0.36  $ & $ -2.56 \pm 1.19  $ & $ -5.99 \pm 0.97  $ \\
ds\_20\_14\_6 & MLP & mlp-li  & $ \textbf{0.23} \pm 0.08  $ & $ \textbf{0.16}  \pm 0.1 $ & $ \textbf{-0.18} \pm 0.3 $ & $ -2.58 \pm 0.95  $ & $ -6.05 \pm 1.38  $ \\
ds\_20\_14\_6 & MLP & xgb-iter  & $ -0.3  \pm 0.43  $ & $ -1.17 \pm 0.73  $ & $ -3.0  \pm 0.9 $ & $ -5.11 \pm 1.21  $ & $ -8.01 \pm 1.82  $ \\
ds\_20\_14\_6 & MLP & xgb-li  & $ -0.31 \pm 0.39  $ & $ -1.05 \pm 0.79  $ & $ -1.96 \pm 0.96  $ & $ -3.79 \pm 1.06  $ & $ -6.26 \pm 1.06  $ \\
\hline
ds\_20\_14\_6 & RF  & knn & $ \textbf{0.83} \pm 0.16  $ & $ \textbf{0.65} \pm 0.3 $ & $ \textbf{0.44} \pm 0.34  $ & $ \textbf{0.04}  \pm 0.41  $ & $ \textbf{-0.6}  \pm 0.78  $ \\
ds\_20\_14\_6 & RF  & linreg-iter & $ \textbf{0.77} \pm 0.1 $ & $ 0.49  \pm 0.3 $ & $ -0.17 \pm 0.59  $ & $ -4.96 \pm 2.12  $ & $ -11.36  \pm 3.15  $ \\
ds\_20\_14\_6 & RF  & linreg-li & $ \textbf{0.73} \pm 0.1  $ & $ 0.45 \pm 0.37  $ & $ -0.61  \pm 1.49  $ & $ -4.22 \pm 2.26  $ & $ -10.34  \pm 4.64  $ \\
ds\_20\_14\_6 & RF  & mice  & $ 0.43  \pm 0.35  $ & $ -0.4  \pm 0.66  $ & $ -2.83 \pm 1.93  $ & $ -6.08 \pm 2.31  $ & $ -10.03  \pm 3.22  $ \\
ds\_20\_14\_6 & RF  & mlp & $ 0.69  \pm 0.15  $ & $ 0.4 \pm 0.3 $ & $ 0.06  \pm 0.38  $ & $ -2.82 \pm 1.72  $ & $ -5.35 \pm 1.13  $ \\
ds\_20\_14\_6 & RF  & mlp-li  & $ \textbf{0.71}  \pm 0.19  $ & $ 0.54  \pm 0.28  $ & $ -0.03 \pm 0.42  $ & $ -2.85 \pm 0.89  $ & $ -5.48 \pm 1.29  $ \\
ds\_20\_14\_6 & RF  & xgb-iter  & $ 0.32  \pm 0.32  $ & $ -0.54 \pm 0.97  $ & $ -2.22 \pm 1.2 $ & $ -4.67 \pm 1.63  $ & $ -7.67 \pm 2.06  $ \\
ds\_20\_14\_6 & RF  & xgb-li  & $ 0.36  \pm 0.4 $ & $ -0.26 \pm 0.65  $ & $ -1.62 \pm 1.05  $ & $ -3.68 \pm 1.58  $ & $ -6.53 \pm 1.49  $ \\
\hline
ds\_50\_35\_15  & MLP & knn & $ \textbf{-0.05} \pm 0.06  $ & $ -0.25 \pm 0.13  $ & $ -0.95 \pm 0.23  $ & $ -2.86 \pm 0.98  $ & $ -7.32 \pm 1.64  $ \\
ds\_50\_35\_15  & MLP & linreg-iter & $ \textbf{-0.03} \pm 0.03  $ & $ \textbf{-0.06} \pm 0.08  $ & $ -0.27 \pm 0.25  $ & $ -2.77 \pm 0.96  $ & $ -6.4  \pm 1.22  $ \\
ds\_50\_35\_15  & MLP & linreg-li & $ \textbf{-0.02}  \pm 0.03  $ & $ \textbf{-0.05}  \pm 0.06  $ & $ \textbf{-0.19}  \pm 0.19  $ & $ -2.62  \pm 0.57  $ & $ -6.77  \pm 0.9  $ \\
ds\_50\_35\_15  & MLP & mice  & $ -0.66 \pm 0.35  $ & $ -1.38 \pm 0.47  $ & $ -1.57 \pm 0.33  $ & $ -2.27 \pm 0.31  $ & $ \textbf{-3.51} \pm 0.62  $ \\
ds\_50\_35\_15  & MLP & mlp & $ \textbf{-0.05} \pm 0.06  $ & $ \textbf{-0.07} \pm 0.06  $ & $ -0.2  \pm 0.12  $ & $ -1.98 \pm 0.6 $ & $ -5.17 \pm 0.92  $ \\
ds\_50\_35\_15  & MLP & mlp-li  & $ \textbf{-0.06} \pm 0.06  $ & $ \textbf{-0.1}  \pm 0.09  $ & $ -0.26 \pm 0.22  $ & $ \textbf{-1.34} \pm 0.47  $ & $ \textbf{-3.82} \pm 0.78  $ \\
ds\_50\_35\_15  & MLP & xgb-iter  & $ -0.7  \pm 0.26  $ & $ -1.75 \pm 0.71  $ & $ -3.41 \pm 0.9 $ & $ -5.63 \pm 0.92  $ & $ -8.16 \pm 1.43  $ \\
ds\_50\_35\_15  & MLP & xgb-li  & $ -0.69 \pm 0.39  $ & $ -1.4  \pm 0.54  $ & $ -2.7  \pm 0.87  $ & $ -4.2  \pm 0.83  $ & $ -6.12 \pm 1.3 $ \\
\hline
ds\_50\_35\_15  & XGBT  & knn & $ \textbf{-0.09} \pm 0.17  $ & $ -0.89 \pm 0.45  $ & $ -2.27 \pm 1.02  $ & $ -5.02 \pm 1.32  $ & $ -9.9  \pm 2.96  $ \\
ds\_50\_35\_15  & XGBT  & linreg-iter & $ \textbf{-0.05} \pm 0.1 $ & $ -0.1  \pm 0.15  $ & $ \textbf{-0.72} \pm 0.46  $ & $ -4.98 \pm 1.44  $ & $ -8.71 \pm 2.36  $ \\
ds\_50\_35\_15  & XGBT  & linreg-li & $ \textbf{-0.02}  \pm 0.08  $ & $ \textbf{-0.08}  \pm 0.19  $ & $ \textbf{-0.46}  \pm 0.39  $ & $ -3.62  \pm 1.32  $ & $ -9.95  \pm 1.76  $ \\
ds\_50\_35\_15  & XGBT  & mice  & $ -0.26 \pm 0.23  $ & $ -0.69 \pm 0.3 $ & $ -0.95 \pm 0.68  $ & $ \textbf{-1.43} \pm 0.47  $ & $ \textbf{-2.65} \pm 0.88  $ \\
ds\_50\_35\_15  & XGBT  & mlp & $ \textbf{-0.02} \pm 0.09  $ & $ \textbf{-0.14} \pm 0.16  $ & $ \textbf{-0.57} \pm 0.37  $ & $ -3.19 \pm 0.86  $ & $ -6.78 \pm 1.36  $ \\
ds\_50\_35\_15  & XGBT  & mlp-li  & $ \textbf{-0.02} \pm 0.16  $ & $ \textbf{-0.09} \pm 0.16  $ & $ \textbf{-0.39} \pm 0.27  $ & $ -2.41 \pm 0.66  $ & $ -5.04 \pm 1.04  $ \\
ds\_50\_35\_15  & XGBT  & xgb-iter  & $ -1.06 \pm 0.77  $ & $ -2.18 \pm 0.96  $ & $ -4.52 \pm 1.66  $ & $ -7.63 \pm 1.81  $ & $ -9.55 \pm 2.45  $ \\
ds\_50\_35\_15  & XGBT  & xgb-li  & $ -0.71 \pm 0.63  $ & $ -1.96 \pm 0.71  $ & $ -3.86 \pm 1.12  $ & $ -5.66 \pm 1.35  $ & $ -8.23 \pm 2.23  $ \\
\hline

\hline
\multicolumn{8}{|c|}{Table continues on the next page.}\\
\hline
\end{tabular}
\end{center}
\end{table}

\begin{table}[t]
\begin{center}
\tiny
\begin{tabular}{| c | c | c | c | c | c | c | c |}
\hline
\multicolumn{8}{|c|}{Continuation of Table \ref{tab:multiple_artificial}.}\\
\hline
Dataset & Model & Method & 10\% & 20\% & 30\% & 40\% & 50\% \\
\hline


ds\_100\_70\_30 & MLP & knn & $ -0.3  \pm 0.16  $ & $ -0.55 \pm 0.19  $ & $ -1.33 \pm 0.2 $ & $ -3.46 \pm 0.9 $ & $ -7.39 \pm 1.21  $ \\
ds\_100\_70\_30 & MLP & linreg-iter & $ \textbf{-0.04} \pm 0.04  $ & $ \textbf{-0.09} \pm 0.08  $ & $ \textbf{-0.28} \pm 0.23  $ & $ -2.41 \pm 0.58  $ & $ -6.83 \pm 1.16  $ \\
ds\_100\_70\_30 & MLP & linreg-li & $ \textbf{-0.02}  \pm 0.04  $ & $ \textbf{-0.05}  \pm 0.06  $ & $ \textbf{-0.1} \pm 0.1 $ & $ \textbf{-0.41}  \pm 0.14  $ & $ \textbf{-1.08}  \pm 0.3 $ \\
ds\_100\_70\_30 & MLP & mice  & $ -0.97 \pm 0.38  $ & $ -1.04 \pm 0.36  $ & $ -2.19 \pm 0.44  $ & $ -3.13 \pm 0.75  $ & $ -3.88 \pm 0.4 $ \\
ds\_100\_70\_30 & MLP & mlp & $ \textbf{-0.09} \pm 0.07  $ & $ \textbf{-0.16} \pm 0.13  $ & $ \textbf{-0.43} \pm 0.21  $ & $ -2.42 \pm 0.52  $ & $ -6.04 \pm 0.86  $ \\
ds\_100\_70\_30 & MLP & mlp-li  & $ -0.19 \pm 0.18  $ & $ -0.31 \pm 0.17  $ & $ \textbf{-0.53} \pm 0.27  $ & $ \textbf{-1.11} \pm 0.3 $ & $ -2.58 \pm 0.77  $ \\
ds\_100\_70\_30 & MLP & xgb-iter  & $ -0.88 \pm 0.26  $ & $ -2.28 \pm 0.63  $ & $ -4.79 \pm 0.91  $ & $ -7.86 \pm 1.23  $ & $ -11.64  \pm 1.13  $ \\
ds\_100\_70\_30 & MLP & xgb-li  & $ -0.88 \pm 0.17  $ & $ -1.8  \pm 0.51  $ & $ -3.4  \pm 0.73  $ & $ -5.43 \pm 0.89  $ & $ -8.22 \pm 0.73  $ \\
\hline
ds\_100\_70\_30 & k-NN  & knn & $ -0.27 \pm 0.25  $ & $ -0.47 \pm 0.29  $ & $ -1.43 \pm 0.59  $ & $ -2.94 \pm 0.67  $ & $ -4.05 \pm 1.21  $ \\
ds\_100\_70\_30 & k-NN  & linreg-iter & $ \textbf{-0.03} \pm 0.13  $ & $ \textbf{0.03}  \pm 0.2 $ & $ \textbf{-0.02} \pm 0.41  $ & $ -1.52 \pm 0.57  $ & $ -4.11 \pm 0.8 $ \\
ds\_100\_70\_30 & k-NN  & linreg-li & $ \textbf{-0.05} \pm 0.16  $ & $ \textbf{0.04} \pm 0.21  $ & $ \textbf{0.01} \pm 0.24  $ & $ \textbf{-0.5} \pm 0.45  $ & $ \textbf{-2.45}  \pm 0.76  $ \\
ds\_100\_70\_30 & k-NN  & mice  & $ -0.75 \pm 0.47  $ & $ -1.23 \pm 0.54  $ & $ -2.15 \pm 0.47  $ & $ -3.2  \pm 0.89  $ & $ -4.07 \pm 0.74  $ \\
ds\_100\_70\_30 & k-NN  & mlp & $ \textbf{0.01} \pm 0.21  $ & $ \textbf{0.01}  \pm 0.2 $ & $ \textbf{-0.24} \pm 0.46  $ & $ -1.24 \pm 0.64  $ & $ -3.12 \pm 0.95  $ \\
ds\_100\_70\_30 & k-NN  & mlp-li  & $ \textbf{-0.03} \pm 0.27  $ & $ \textbf{0.0} \pm 0.32  $ & $ \textbf{-0.15} \pm 0.3 $ & $ \textbf{-0.69} \pm 0.5 $ & $ \textbf{-2.0}  \pm 0.58  $ \\
ds\_100\_70\_30 & k-NN  & xgb-iter  & $ -0.93 \pm 0.67  $ & $ -2.13 \pm 0.93  $ & $ -4.16 \pm 1.11  $ & $ -6.28 \pm 1.68  $ & $ -8.71 \pm 1.28  $ \\
ds\_100\_70\_30 & k-NN  & xgb-li  & $ -1.21 \pm 0.56  $ & $ -2.06 \pm 0.7 $ & $ -3.55 \pm 1.27  $ & $ -5.25 \pm 0.89  $ & $ -7.23 \pm 1.29  $ \\
\hline

\end{tabular}
\end{center}
\end{table}

\bibliography{arxiv}{}
\bibliographystyle{unsrt}


\end{document}